\documentclass[lettersize]{article}
\usepackage{latexsym}

\usepackage[bf]{caption}
\usepackage{graphicx, subfig, times,amsmath, txfonts}
\usepackage[square,numbers,sort&compress]{natbib}
\usepackage[pdfborder={0 0 0}]{hyperref}

\renewcommand{\abstract}[1]{\begin{quote}\small \noindent #1 \end{quote} }

\newcommand{\keywords}[1]{\begin{quote}\small \noindent {\bf Keywords } #1 \end{quote} }

\graphicspath{{./images/}}

\title{\bf Exploring Promising Stepping Stones by Combining \\ Novelty Search with Interactive Evolution}  

\author{
	Brian G. Woolley and Kenneth O. Stanley \\
	\small{brian.woolley@ieee.org, kstanley@eecs.ucf.edu} \\
	\small{Department of Electrical Engineering and Computer Science} \\
	\small{University of Central Florida, Orlando, Florida 32816, USA}
}

\date{\small{\today}}

\begin{document}

\maketitle

\clubpenalty=10000
\widowpenalty = 10000

\abstract{The field of evolutionary computation is inspired by the achievements of natural evolution, in which there is no final objective. Yet the pursuit of objectives is ubiquitous in simulated evolution. A significant problem is that objective approaches assume that intermediate stepping stones will increasingly resemble the final objective when in fact they often do not. The consequence is that while solutions may exist, searching for such objectives may not discover them. This paper highlights the importance of leveraging human insight \emph{during} search as an alternative to articulating explicit objectives.  In particular, a new approach called \emph{novelty-assisted interactive evolutionary computation} (NA-IEC) combines human intuition with novelty search for the first time to facilitate the serendipitous discovery of agent behaviors. In this approach, the human user directs evolution by selecting what is interesting from the on-screen population of behaviors. However, unlike in typical IEC, the user can now request that the next generation be filled with \emph{novel} descendants.  The experimental results demonstrate that combining human insight  with novelty search finds solutions significantly faster and at lower genomic complexities than fully-automated processes, including pure novelty search, suggesting an important role for human users in the search for solutions.}


%-------------------------------------
\section{Introduction}\label{sec:intro}
%-------------------------------------
Several results in recent years have hinted at the limitations of traditional objective functions, wherein the more a candidate resembles the objective, the higher its fitness.   Whether the objective is to evolve a particular behavior like balancing a pole~\citep{gruau:gp96, stanley:ec02} or a particular morphology like a French flag via a developmental system~\citep{miller:gecco04}, such objective-based evolution is the dominant approach across a wide breadth of domains and methods.  An early hint that such an approach to fitness may be flawed was from experiments with the \emph{novelty search} algorithm~\citep{lehman:alife08}, which rewards novel behaviors instead of rewarding objective performance.  Interestingly, novelty search significantly outperformed objective-based fitness in a deceptive maze-navigation domain~\citep{lehman:alife08, lehman:ecj11}, showing counter-intuitively that in some deceptive cases it is possible that having no specific objective may work better than rewarding progress toward the objective.
 
Yet novelty search is not the only hint that something is amiss with fitness.   A second hint was from \citet{woolley:gecco11}, who studied what happens when an attempt is made to \emph{re-evolve} images that were previously evolved interactively by human users on the Picbreeder online service~\citep{secretan:ecj11}.  The strange result was that none of the more interesting images, such as the \emph{Butterfly} and the \emph{Skull}, could be re-evolved by the very same evolutionary algorithm when they are made the automated objective.   In other words, even though a set of users together evolved a picture of a \emph{Skull} in only 74 generations, 20 automated attempts of 30,000 generations each were unable to reproduce the result.  Picbreeder is full of such images, i.e.~each evolved by users in just a few dozen generations and with no specific objectives, yet each nearly impossible to reproduce when they are made objectives.
 
While such results are intriguing, their interpretation has also been controversial.  Although they suggest that searches not driven by explicit objectives might sometimes offer more potential than those that are, they seem to offer few alternatives other than searching only for novelty or leaving the search entirely to human guidance.  However, work with novelty search has shown that it may become lost in especially large spaces~\citep{lehman:gecco10a, kistemaker:gecco11}, and \citet{takagi:ieee01} warns that interactive evolutionary computation (IEC) is limited by human fatigue.  With such limitations for alternative approaches, the news that traditional objectives offer little hope is not especially encouraging.
 
One potential response is to hybridize an objective-based search with a search for novelty, as in Novelty-Based Multiobjectivization~\citep{mouret:iros11}.  Yet while this idea undoubtedly works in some cases, in others the reintroduction of the objective, even partially, only disadvantages the search.  After all, as recent critiques of objective-based search have pointed out, the fundamental problem with objectives is that they often penalize essential intermediate stepping stones that lead to the objective because those stepping stones do not resemble the objective~\citep{ficici:alife98, lehman:alife08, lehman:ecj11, woolley:gecco11}.  This problem, called \emph{deception}, by definition is exacerbated by reintroducing a deceptive objective back into the search. 

Such concerns do not imply that objectives are never useful, or that hybrid objective/non-objective approaches cannot help; rather they open the door to the possibility that more can be done to emphasize the discovery of essential stepping stones.   For example, Picbreeder suggests that humans are uniquely adept at identifying promising stepping stones, even if their ultimate destination is entirely unclear~\citep{woolley:gecco11}.  Such serendipitous exploration of a large search space is particularly attractive in fields like generative and development systems (GDS)~\citep{stanley:alife03, hornby:cec01, bongard:cec02}, where often the deeper motivation behind experiments is to demonstrate the power of the encoding as opposed to evolving a particular artifact (i.e.~the field of GDS is not \emph{inherently} interested in French flags). 
 
Thus the main insight in this paper is that the ability of humans to identify promising stepping stones is naturally complemented by the ability of novelty search to generate \emph{candidate sets} of potential stepping stones.  In other words, novelty search can mitigate the main weakness of IEC (i.e.~that humans grow tired quickly~\citep{takagi:ieee01}) by offloading most of the exploratory work.  This way, novelty search becomes a kind of stepping-stone scavenger that is interleaved with human evaluations that determine which stepping stones are the most promising.  Furthermore, neither the human nor the novelty search are guided by any explicit objective, thereby also mitigating the threat of deception.  In this approach, instead of forcing a human experimenter to \emph{articulate} through a fitness function exactly what should be rewarded in a complex domain, the human instead can leverage highly-nuanced \emph{implicit} hunches that all of us have about what is promising.  The result is a powerful synergy between two promising non-objective processes that reintroduces to novelty search a sense of control (i.e.~from the human) without reintroducing an explicit objective.
 
In this paper, this approach, called \emph{novelty-assisted IEC} (NA-IEC), is compared to pure novelty search and objective-based search in evolving neurocontrollers for robots in the deceptive mazes of \citet{lehman:ecj11}.  Interestingly, while novelty search was previously shown significantly more effective than objective-based search in this domain~\citep{lehman:ecj11}, NA-IEC outperforms novelty search by a multiple of three to four times, yielding by far the fastest solution on these deceptive problems.  Furthermore, NA-IEC is also eight to ten times faster in \emph{clock time}, even with the human in the loop, suggesting that perhaps the effort spent crafting objectives functions, which are often deceptive anyway, would be better spent in obtaining a small number of suggestions from a human evaluator during the search process itself.

%-------------------------------------
\section{Background}\label{sec:background}
%-------------------------------------
This section reviews deception in EC and the non-objective methods that are the basis for the approach introduced in this paper.

%-------------------------------------
\subsection{Deceptive Task Domains}\label{sec:mazeDomain}
%-------------------------------------
The key question in research on deception is what causes evolutionary algorithms (EAs) to fail and how to mitigate such failures~\citep{whitley:fga91, goldberg:gabook89}.  For the purpose of this work, we are interested in the case in which pursuing what appears to be a reasonable \emph{objective} produces an unreasonable \emph{objective function}.  In this context, an intuitive definition of deception, as stated by \citet{lehman:ecj11}, is:  ``A deceptive objective function will \emph{deceive} search by actively pointing the wrong way.''

% Introduce Deceptive/Rugged Fitness Lanscapes
The fitness function can point the wrong way because not only must it reward the objective, but it must also reward the intermediate solutions (i.e.~the \emph{stepping stones}) that lead to the objective.  Often these stepping stones do not improve performance on the objective function (and may even decrease it), causing search algorithms to forsake the most promising candidates.

\begin{figure} [!t]
\centering
	\vspace{-1em}
	\subfloat[Medium Map]{
		\label{fig:medMap}
		\includegraphics[width=0.338\linewidth]{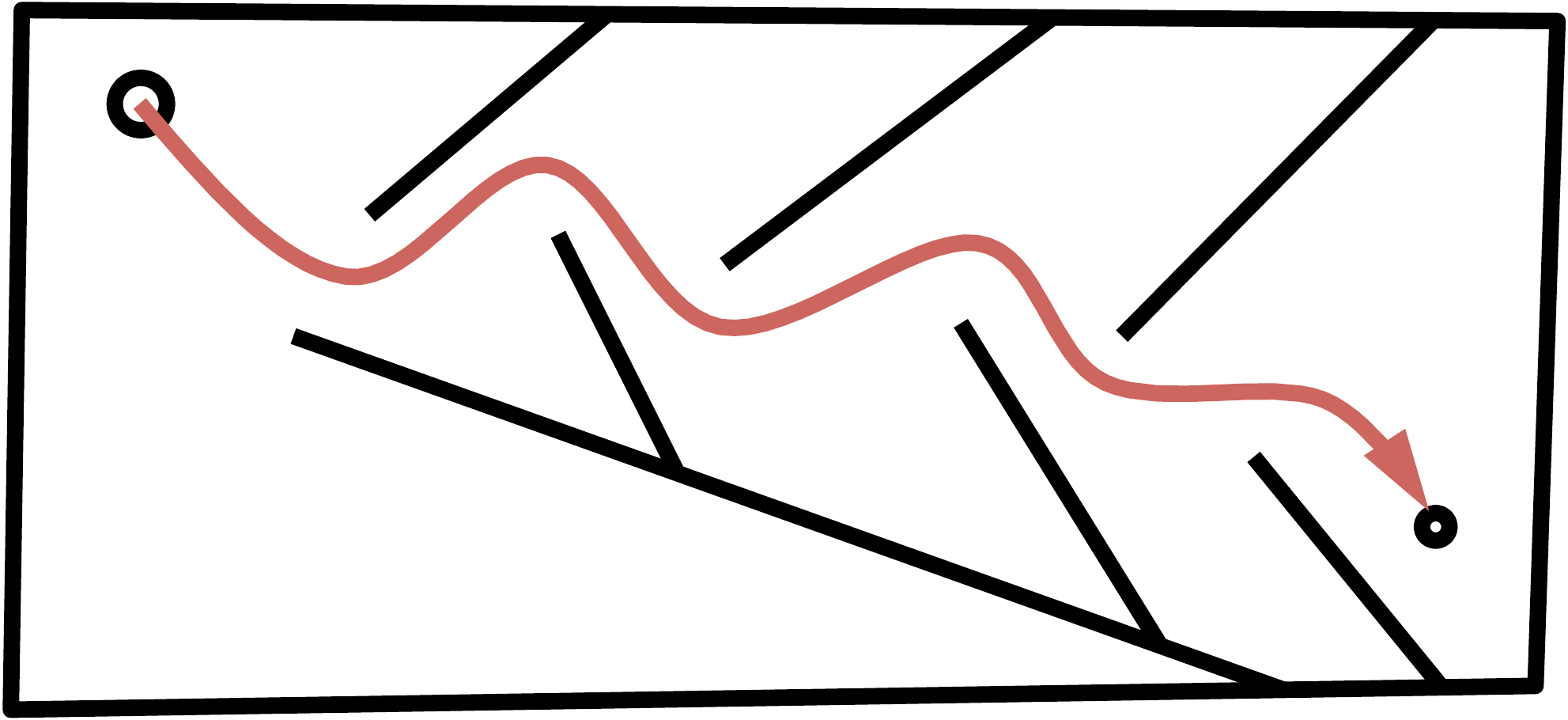}
	}
	\hspace{5em}
	\subfloat[Hard Map]{
		\label{fig:hardMap}
		\includegraphics[width=0.225\linewidth]{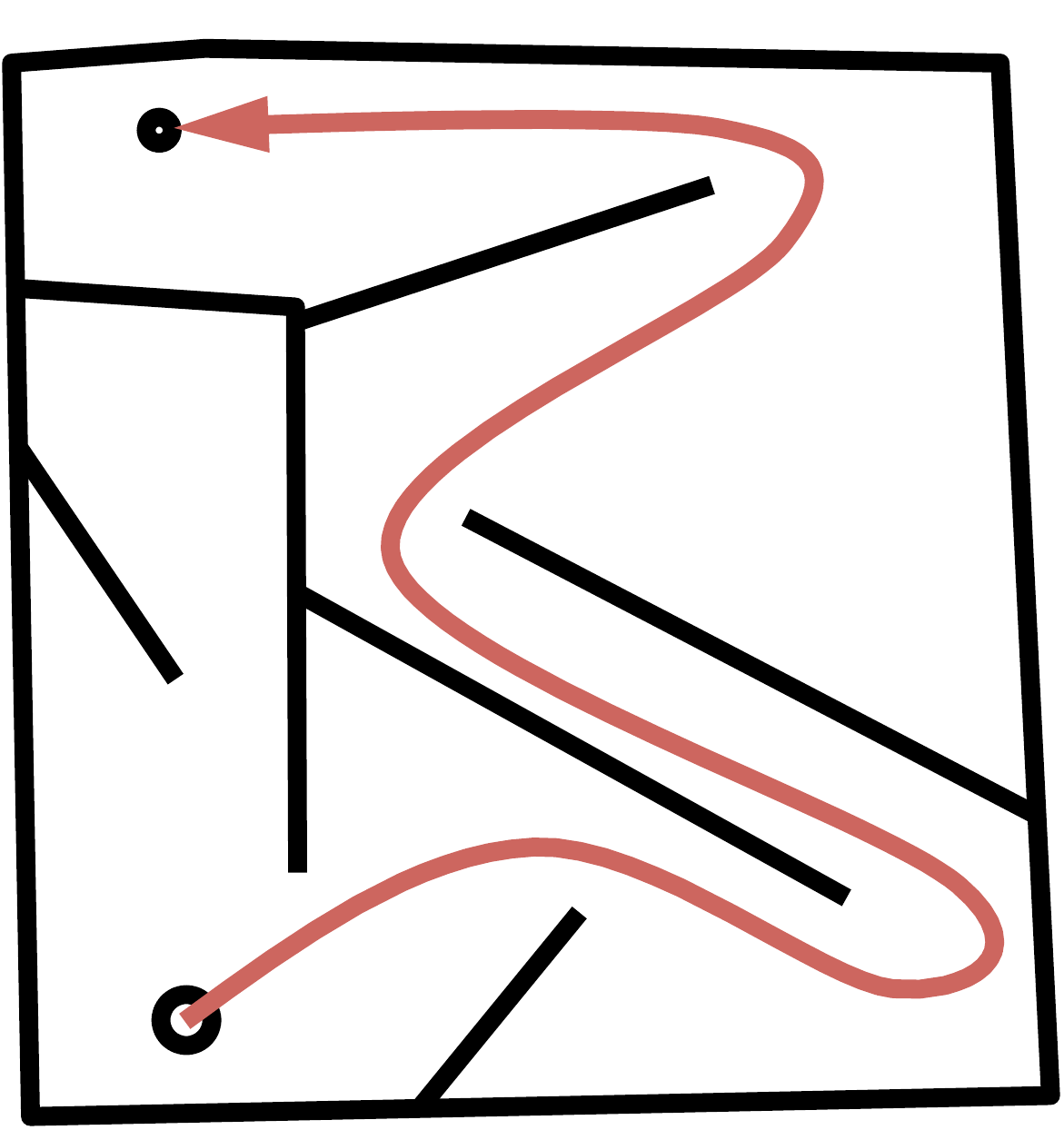}
	}
	\caption{\textbf{Maze Navigation Maps (\citet{lehman:alife08, lehman:ecj11}).}  The deceptive maze domain is a metaphor for search and is \emph{not} a path-planning problem.  Rather, the aim is to evolve a neural network that drives a robot through the maze.  The walls represent barriers to search and the cul-de-sacs represent local-optima that can deceive objective-based search.}
	\label{fig:maps}
\end{figure}

\begin{figure} [!t]
\centering
	\hspace{-3em}
	\subfloat[Sensors]{
		\label{fig:robotSensors}
		\includegraphics[width=0.375\linewidth]{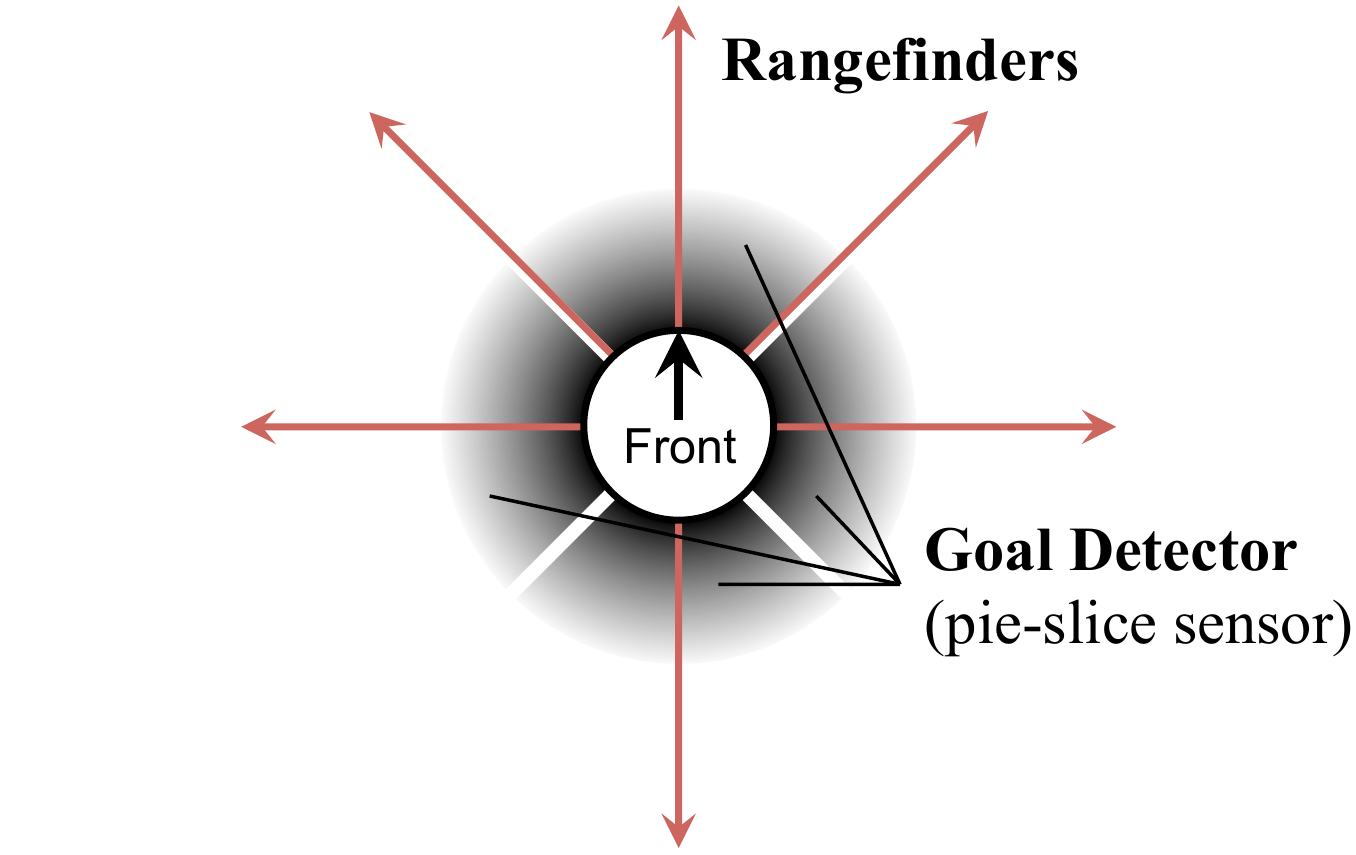} %0.4
	}
	\hspace{3em}
	\subfloat[Neural Network]{
		\label{fig:robotANN}
		\includegraphics[width=0.375\linewidth]{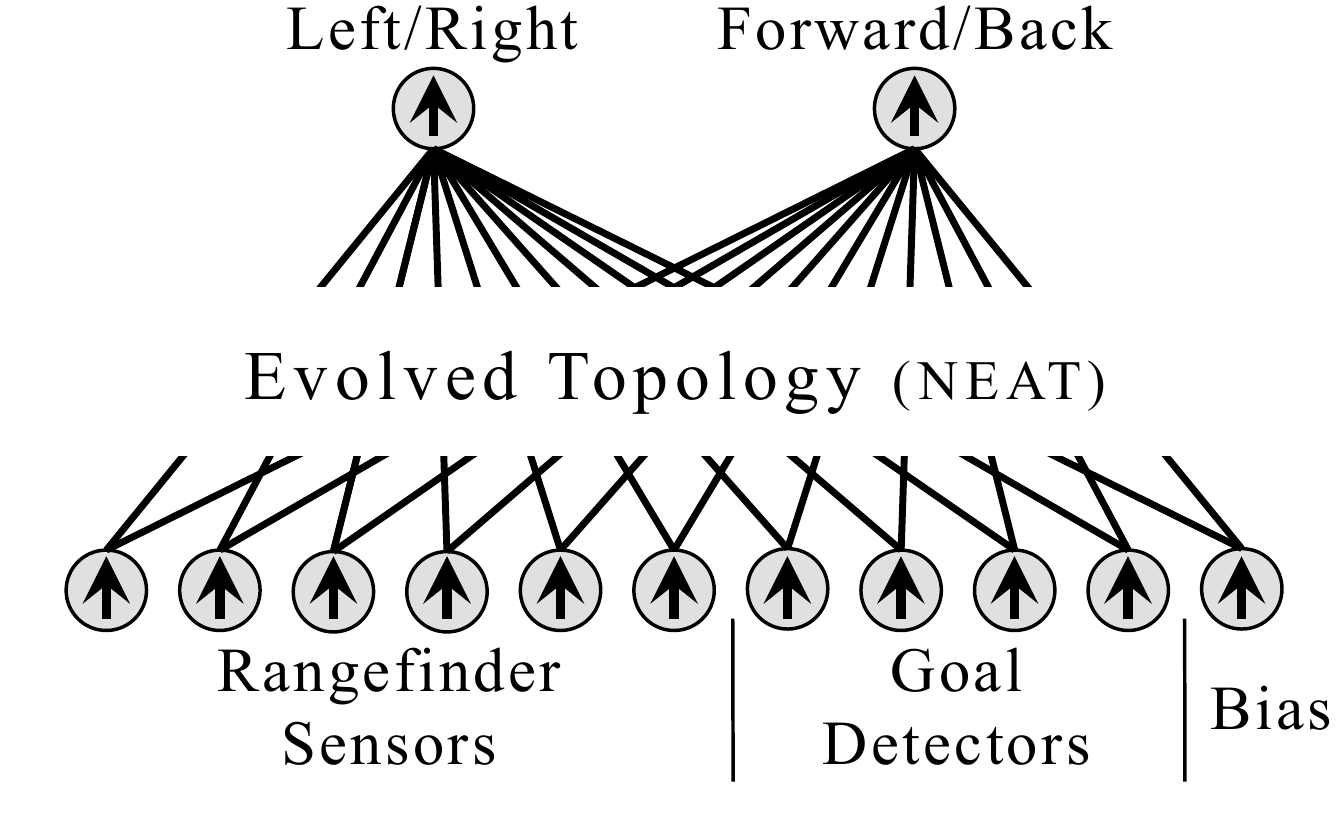} %0.3375
	}
	\caption{\textbf{Maze Navigation Robot (\citet{lehman:alife08, lehman:ecj11}).}  The sensor package (a) includes six rangefinders that detect walls and four pie-slice sensors that signal the general direction to the objective.  The navigation behavior, encoded as an ANN~(b), maps sensor inputs to actions, i.e.~turn rate (\emph{left}/\emph{right}) and velocity (\emph{forward}/\emph{backward}).  Under this construction, navigators cannot see the whole maze and must evolve a control policy that traverses the maze based on sensory input.}
	\label{fig:robot}
\end{figure}

A good example of a deceptive domain, which is also the experimental domain in this paper, is the deceptive maze domain introduced by \citet{lehman:alife08, lehman:ecj11}, in which a simulated robot must navigate through a maze with deceptive cul-de-sacs (figure~\ref{fig:maps}).  The maze-navigation agents that act within the maze have a sensor package with six rangefinders that detect the walls and four pie-slice sensors that signal the direction to the goal~(figure~\ref{fig:robotSensors}).  Each robot's navigation behavior, encoded as an artificial neural network (ANN), maps sensor inputs to actions, i.e.~turn rate (\emph{left}/\emph{right}) and velocity (\emph{forward}/\emph{backward}), as shown in figure~\ref{fig:robotANN}.  Under this construction, navigators must evolve a control policy that traverses the maze based on sensory input.

The medium and hard maps in figure~\ref{fig:maps} are deceptive by design because the maps contain cul-de-sacs that represent local optima in the search space.  If fitness is assigned based on reducing the distance to the goal, then the objective function prunes out of the search the deceptive intermediate solutions (i.e.~those that move away from the goal location) needed to reach the global objective.  While an alternative objective function that rewards specific intermediate solutions is conceivable (and in fact will be explored later in this paper as well), the original point of this domain was to explore the effect of objective fitness when the precise stepping stones are \emph{not known}, which is the typical predicament in most domains of interest.  In such cases, as in the standard objective function here, performance is generally rewarded for its proximity to the target behavior.  Thus evolution driven by proximity to the goal often converges to a cul-de-sac from which the goal is inaccessible.

Despite extensive research, deception remains a significant problem in the field of EC~\citep{goldberg:icga87, pelikan:traps, liepins:deceptiveness}.  The problem is that evolutionary algorithms (EAs) ultimately respond to the selection pressures created by the fitness function, which is often misleading.  The challenge is to determine how to reward the intermediate steps that are required to reach the goal.  In effect, what appears to be a reasonable heuristic may actually \emph{prevent} the objective from being reached.  Therefore, any similarity metric that guides the search toward an a~priori objective is potentially a false compass to the optimal solution~\citep{stanley:splash10}.

%Transition to Novelty search
In this spirit, \citet{lehman:alife08,lehman:ecj11} introduced the idea of abandoning objectives as a search heuristic in deceptive domains, electing instead to reward individuals only for novel behaviors, as described next.

%-------------------------------------
\subsection{Novelty Search}
%-------------------------------------
A fundamental dilemma with objectives in EC is that defining an effective fitness function is akin to understanding the fitness landscape or knowing the stepping stones a~priori~\citep{ficici:alife98, zaera:sab96}.  Such a requirement becomes increasingly difficult as objectives become more ambitious because the intermediate steps to the solution are less likely to be known~\citep{ficici:alife98}.  As an alternative, \citet{lehman:alife08, lehman:ecj11} demonstrated that searching without regard to the objective, i.e.~searching only for novel behavior, is more effective at discovering solutions in some deceptive domains than rewarding objective performance.

Novelty search works with EAs by replacing the fitness function with a \emph{novelty metric}.  The novelty metric is a measure of the uniqueness of an individual's behavior at a given task.  Instead of rewarding performance, novelty search rewards individuals in the population for finding new ways to complete the evaluation task, thus creating a constant pressure to do something new~\citep{lehman:ecj11}.

Because novelty search operates in \emph{behavior space}, it is important first to characterize the space of unique behaviors in a way that is meaningful to the domain.  The novelty search algorithm then computes the \emph{sparseness} in the behavior space as the average distance to the $k$-nearest neighbors~\citep{cover:book91} around that behavior.  The sparseness $\rho$ of behavior $x$ is given by
\vspace{-1em}
\begin{equation}
	\rho(x)=\frac{1}{k}\sum_{i=0}^{k}{\text{dist}(x,\mu_i)},
	\label{eq:noveltyMetric}
\end{equation}
where $\mu_i$ is the $i$th-nearest neighbor of $x$ with respect to the distance function $\text{dist}(x,\mu)$.  In this way, if the average distance is large, then the candidate solution is considered to be in a sparse area of the behavior space, thus making it more likely to be selected by the EA.  Optionally, as in coevolution~\citep{dejong:gecco04b}, an archive of past behaviors may serve to avoid backtracking through the behavior space.  If the novelty metric is sufficiently high for a new individual (i.e.~above some minimal threshold $\rho_{\text{min}}$), then the individual may be recorded in the permanent archive to provide a comprehensive sample of where the search has been, thereby increasing the pressure to discover new ways of behaving in the domain~\citep{lehman:alife08, lehman:ecj11}.

Characterizing behaviors so that they can be compared is the most challenging aspect of novelty search.  For the deceptive maze domain~\citep{lehman:alife08, lehman:ecj11} (figure~\ref{fig:maps}), the behavior of a maze navigation robot is usually defined as its \emph{final position}.  In this way, the novelty metric rewards controllers that end at new locations in the maze.  At first, the collection of behaviors may include robots that do nothing, get stuck in corners, run in circles, and so forth.  However, at some point, the collection of simple behaviors becomes saturated and the pressure to do something new increases, i.e.~evolution favors mutations that take the navigator to new places in the maze.

While the idea of selecting anything novel may sound potentially similar to exhaustive search, searching in the space of \emph{behaviors} is often tractable because many points in the space of possible \emph{genomes} collapse to a single behavior.  Furthermore, when applied in conjunction with complexifying algorithms like NEAT~\citep{lehman:alife08, lehman:gecco10a} and GP~\citep{lehman:gecco10b}, simple behaviors become associated with minimal representations, and only mutations that increase the size of the genome and lead to novel behaviors are explored further.  Therefore, this approach, operating without regard to an objective, moves into complex spaces in a meaningful way because new behaviors are those that could not be expressed at lower levels of complexity~\citep{ventrella:ms94}, i.e.~complexity is rewarded when it is warranted.

% Deficiency of novelty search and transition to IEC and Picbreeder
However, experience has also shown that novelty search becomes lost in unrestricted domains~\citep{lehman:gecco10a}.  In such domains there is an opportunity to leverage human knowledge rather than exhaustively exploring the space of all possible solutions.  For example, in the space of all possible images, humans recognize the importance of symmetry in pictures and are able to relate structural innovations with objects in the real world.  Thus the next section provides relevant background on the field of human-led evolution, followed by a description of Picbreeder, a domain in which a community of users interactively evolves a collection of meaningful images without having a formal, overall objective.  The NA-IEC approach introduced in this paper will unify such human-led evolution with both novelty and objective-based search.

%-------------------------------------
\subsection{Interactive Evolutionary Computation}
%-------------------------------------
In \emph{interactive evolutionary computation}~(IEC) the traditional obj\-ective-function is replaced by a user who performs selection~\citep{takagi:ieee01}.  IEC is effective in creative domains~\citep{romero:book07} where the term \emph{fitness} is subjective because what people experience as pleasing or interesting is based on individual preferences.  Thus when what is good, bad, meaningful, or strange is too broad and complex to encode into a traditional objective function, interactive evolution can provide a means for making significant discoveries in evolutionary systems.

Like traditional EAs, IEC systems also typically begin from a random initial population that evolves over generations by selecting, mating and mutating members.  However, IEC differs from traditional automated EAs in that a human user is now responsible for the evaluation and selection of promising candidate solutions.  While this difference typically leads to smaller population sizes and higher mutation rates, the most profound implication is that evolution is no longer bound to a rigid expression of what is fit and unfit.  In fact, the human evaluator's breadth of experience makes it likely that his or her selection criteria will change over the course of evolution.  Such an ability to make \emph{serendipitous discoveries}, i.e.~to identify and pursue important artifacts as they emerge, is the primary motivation of the NA-IEC approach introduced in this paper.

To interface with the human evaluator, the majority of IEC systems are modeled after the original Blind Watchmaker \emph{Biomorphs} application by Dawkins~\citep{dawkins:book86}.  In this approach the user is presented with a panel of individuals (e.g.~$3\times4$) from which the parents of the next generation are selected.  The IEC system then mates, recombines, and mutates the genetic material of the parents to create the next generation, which is then presented to the user.  This process is repeated at the user's direction until the user is satisfied.

%The results of Dawkins' original nine-gene Biomorphs~\citep{dawkins:book86} demonstrate that selecting for phenotypic effects, i.e.~selecting for how genes are expressed, leads to meaningful discoveries.  This work was furthered by Sims~\citep{sims:siggraph91}, who interactively evolved variable-length expressions for lifelike three-dimensional plant structures.  In both cases, the insight is that knowing the underlying genetic encoding is not important.  Rather, selecting meaningful phenotype attributes that emerge during evolution does lead to significant results in the vast space of what is possible.

Despite the benefits of having a human in the loop, such IEC systems are limited by user fatigue.  According to Takagi~\citep{takagi:ieee01}, typical IEC only lasts 10--20 generations per session.  The problem is that the vast majority of significant discoveries exist beyond the reach of a single-user session.  One response, which has become known as \emph{collaborative interactive evolution} (CIE~\citep{szumlanski:ncai06}), is to leverage the efforts of many users.  One particularly successful CIE system, and the one that in part inspired the NA-IEC approach, is the Picbreeder project~\citep{secretan:chi08, secretan:ecj11}, which is described next.

%-------------------------------------
\subsection{Picbreeder}
\begin{figure}[!t]
%	\centering
		\hspace{-0.575em}
		\subfloat[Butterfly]{\label{fig:target4376}
			\href{http://picbreeder.org/search/showgenome.php?sid=4376}
			{\includegraphics[width=0.120\linewidth]{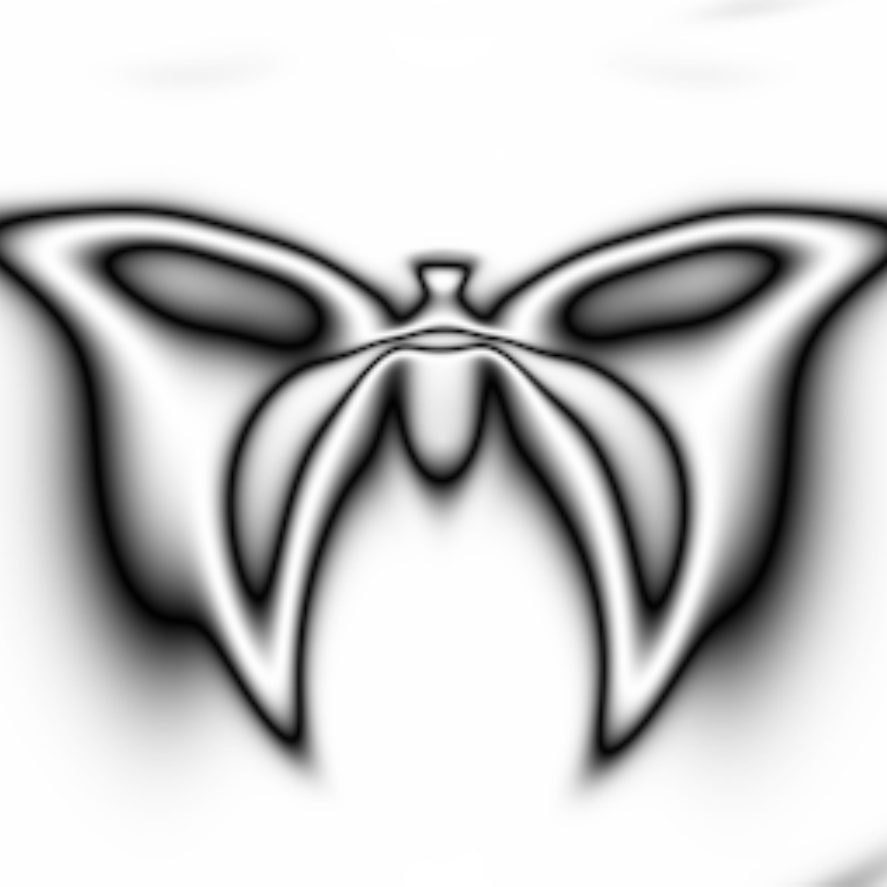}}}
%			\hspace{0.04 in}
		\subfloat[Skull]{\label{fig:target576}
			\href{http://picbreeder.org/search/showgenome.php?sid=576}
			{\includegraphics[width=0.120\linewidth]{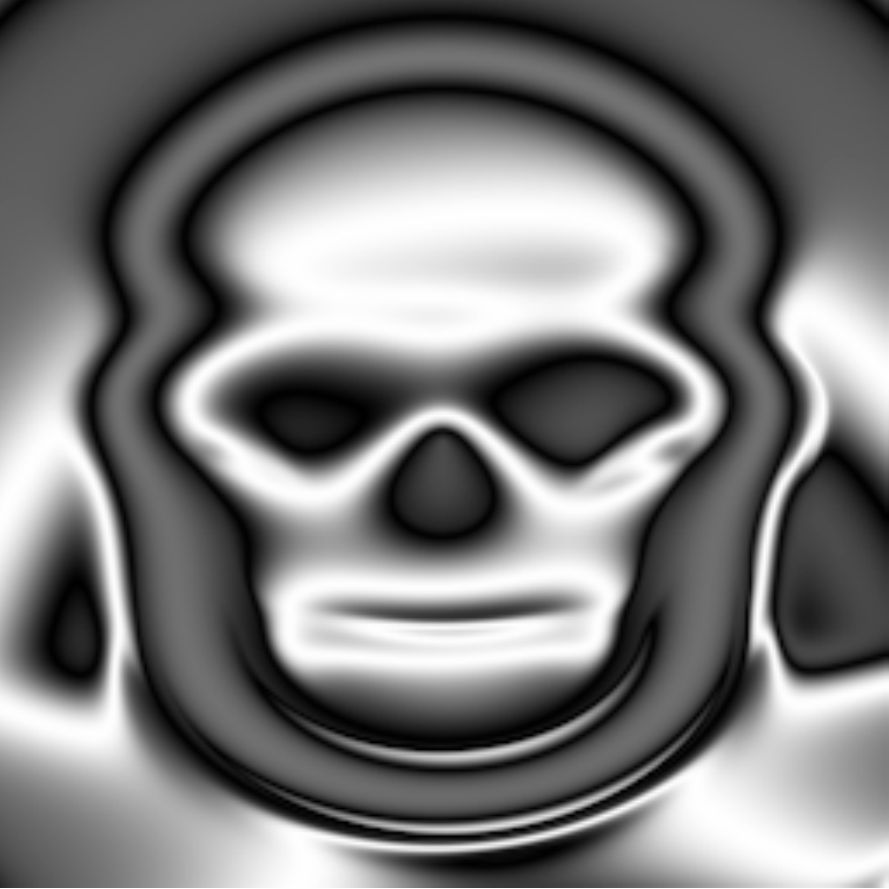}}}
%			\hspace{0.04 in}
		\subfloat[Sunset]{\label{fig:target3817}
			\href{http://picbreeder.org/search/showgenome.php?sid=3817}
			{\includegraphics[width=0.120\linewidth]{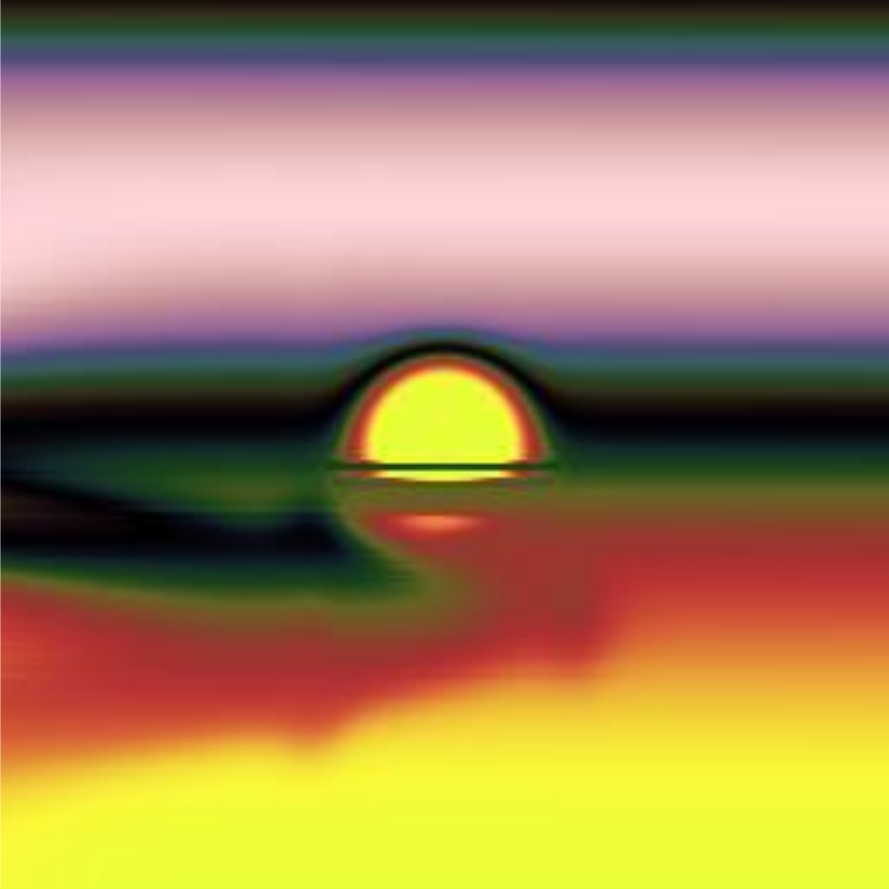}}}
%			\hspace{0.04 in}
		\subfloat[Dolphin]{\label{fig:target4041}
			\href{http://picbreeder.org/search/showgenome.php?sid=4041}
			{\includegraphics[width=0.120\linewidth]{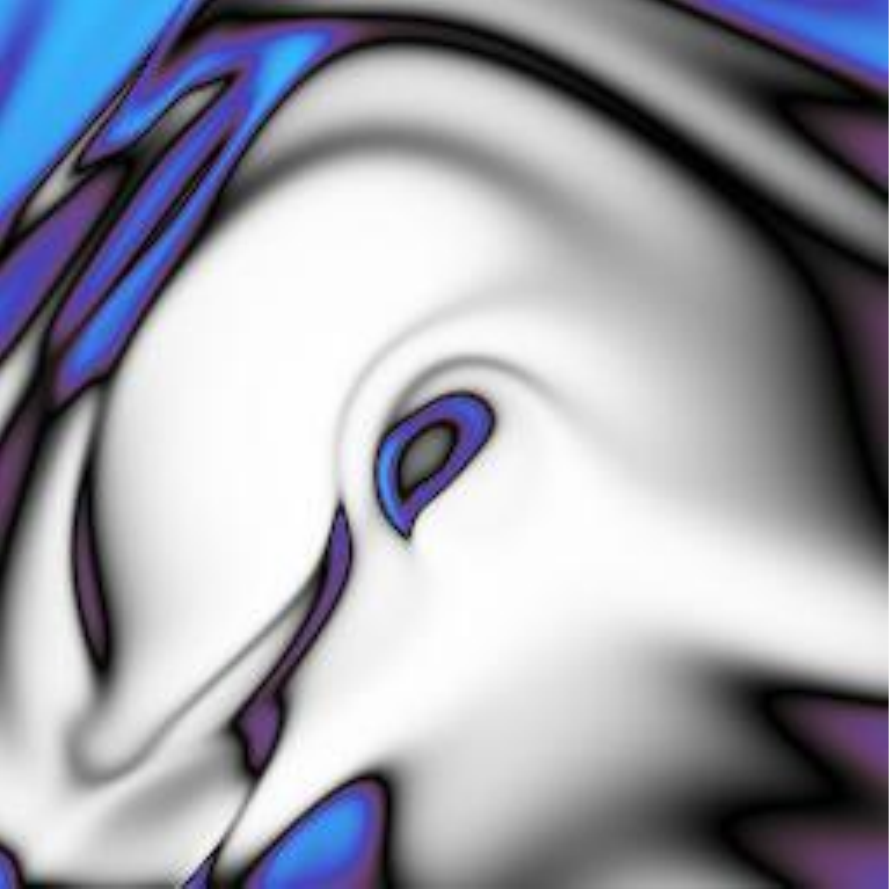}}}
%			\hspace{0.04 in}
		\subfloat[Car]{\label{fig:target3806}
			\href{http://picbreeder.org/search/showgenome.php?sid=3806}
			{\includegraphics[width=0.120\linewidth]{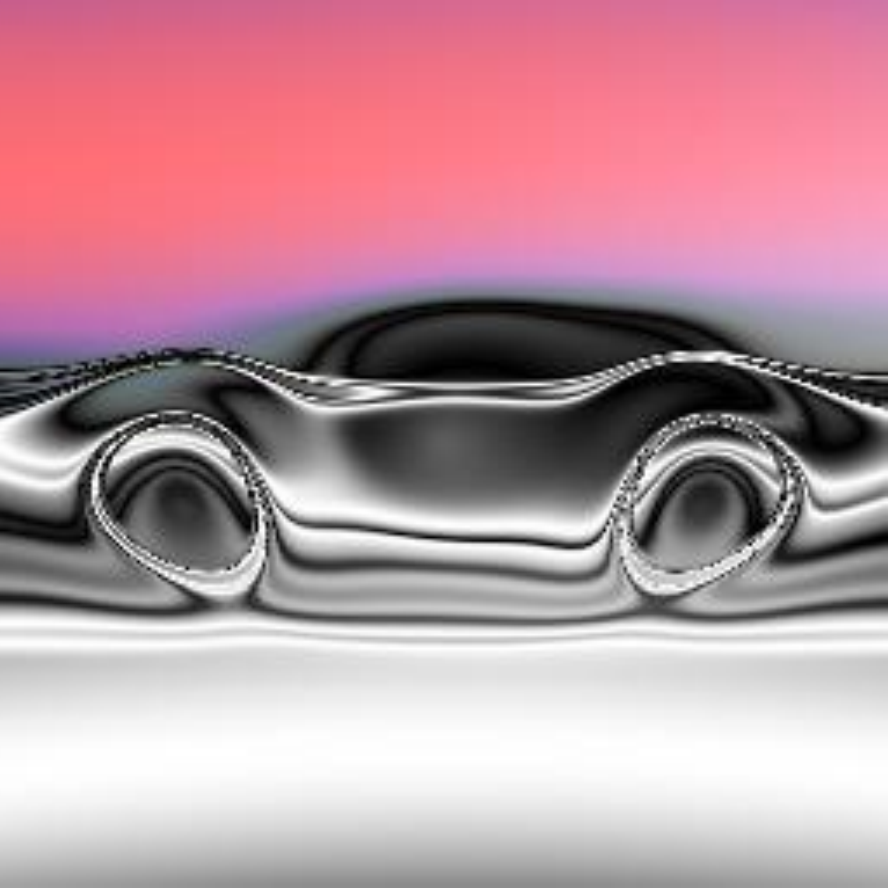}}}
%			\hspace{0.04 in}
		\subfloat[Mystic]{\label{fig:target3674}
			\href{http://picbreeder.org/search/showgenome.php?sid=3952}
			{\includegraphics[width=0.120\linewidth]{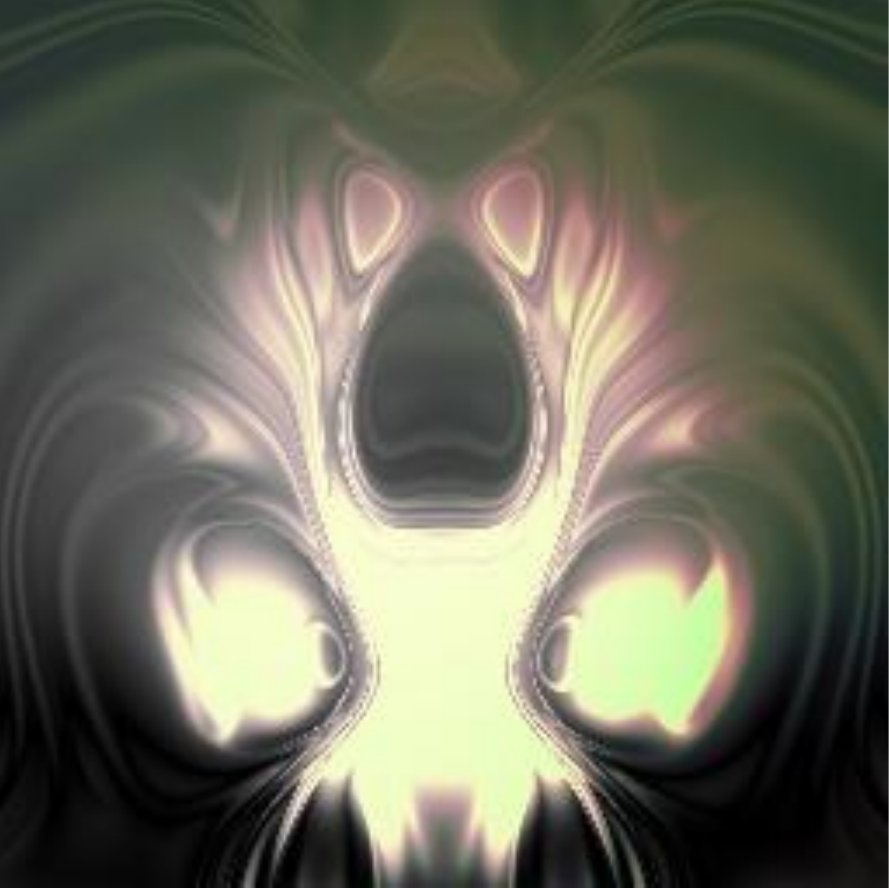}}}
%			\hspace{0.04 in}
		\subfloat[Apple]{\label{fig:target5736}
			\href{http://picbreeder.org/search/showgenome.php?sid=5736}
			{\includegraphics[width=0.120\linewidth]{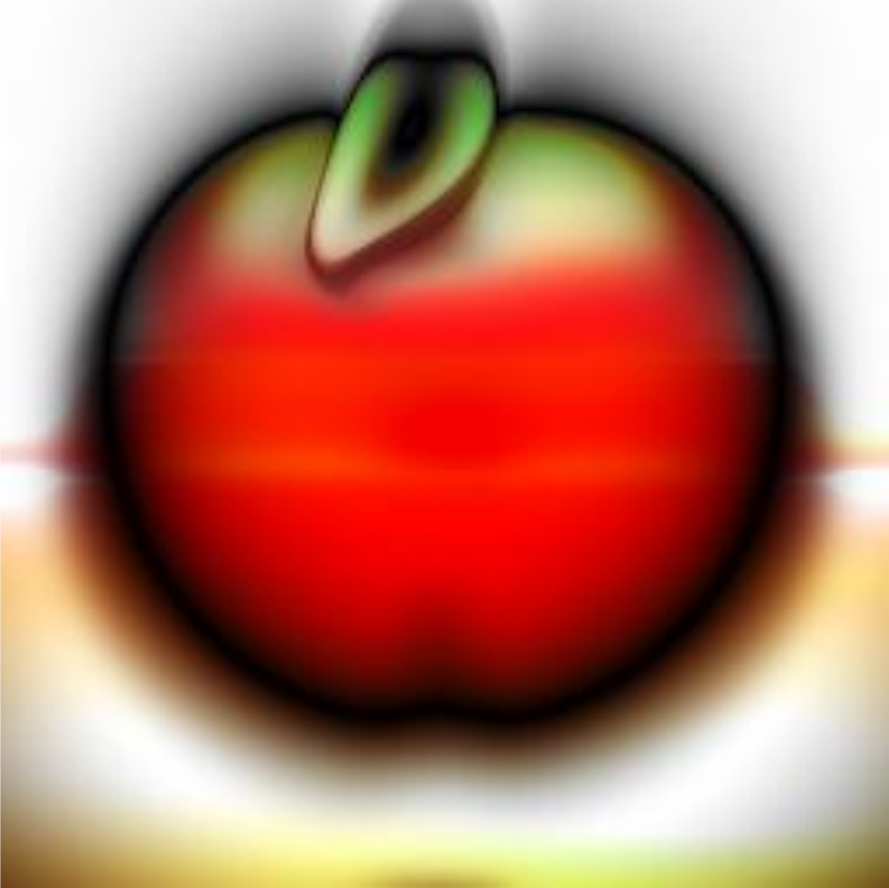}}}
%			\hspace{0.04 in}
		\subfloat[Wizard]{\label{fig:target7506}
			\href{http://picbreeder.org/search/showgenome.php?sid=7506}
			{\includegraphics[width=0.120\linewidth]{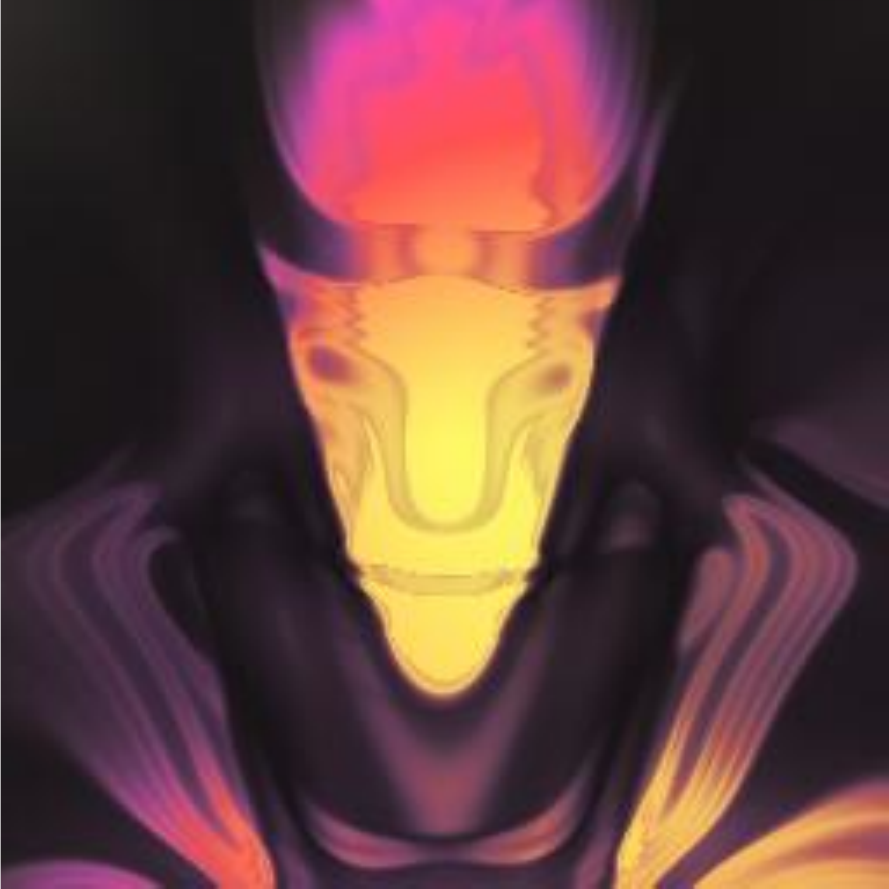}}}
	\caption{\textbf{Images Evolved on Picbreeder~(\citet{secretan:ecj11}).}  These images were interactively evolved by a community of human users with no explicit objective.  They demonstrate the system's ability to discover interesting and meaningful images, including such seminal images as the \emph{Butterfly} and the \emph{Skull}---which were evolved in just 90 and 74 generations, respectively.}
	\label{fig:images}
\end{figure}

%-------------------------------------
\emph{Picbreeder}~({\small\url{http://picbreeder.org}}~\citep{secretan:chi08, secretan:ecj11}) is a distributed community of online users that interactively evolve pictures by selecting images that are appealing.  Picbreeder is a CIE system because users on Picbreeder \emph{collaborate} by continuing to evolve images previously evolved by other users.  The collection of images generated by Picbreeder is significant because it demonstrates how a group of individuals working without a formal unified objective can discover \emph{attractive} and \emph{interesting} areas in the vast desert of all possible images; some such images are shown in figure~\ref{fig:images}.  Additionally, the quality of such a serendipitous approach to evolution is evident in the diverse phylogeny of images that have emerged, the compactness of their representations, and the speed (i.e.~low number of generations) with which meaningful images are discovered.

%  Images from Picbreeder.  

Users evolve images in Picbreeder by selecting ones that appeal to them from among a set of 15 candidates to produce a new generation.  As this process is repeated, the individual images in the population evolve to satisfy the user.  Once satisfied, the user can \emph{publish} his or her image to the Picbreeder site.  Sharing their work with the community then allows others to continue evolving already-published images to form new and more intricate designs~\citep{secretan:chi08}, which is called \emph{branching}.

% Transition to introducing NA-IEC
In this way, interactive evolution can discover meaningful artifacts that were not known to exist a priori in the space of all possible images; some such images are shown in figure~\ref{fig:images}.  Interestingly, \citet{woolley:gecco11} showed that human discoveries in Picbreeder like the \emph{Skull} and \emph{Butterfly} cannot be rediscovered by the very same algorithm as in Picbreeder (NEAT;~\citep{stanley:ec02,stanley:jair04}) when evolving such images is made the explicit objective in an automated evolution.  This result hints at the potential for missed opportunities when objective-based search is deployed on its own.  Inspired by both novelty search and Picbreeder, the next section introduces a new evolutionary framework in which human users influence not only the direction, but also the mode of evolution.

%-------------------------------------
\section{NA-IEC Framework} \label{sec:naiec}
%-------------------------------------
The main idea in this section is to combine for the first time the intuitive ability of human users to identify what is interesting and important in a domain, i.e.~interactive evolutionary computation (IEC), with a stepping stone generator based on a short-term novelty search and an objective optimizer to create a synergistic effect that expedites the evolution of controller solutions.  Under this new approach, called \emph{Novelty-Assisted Interactive Evolutionary Computation} (NA-IEC), a human user is asked to select individuals from a population of candidate behaviors and then apply one of three evolutionary operations:  a traditional IEC step, a short-term novelty search, or a fitness-based optimization.

In this way, the user can apply the evolutionary operations where appropriate, even changing the mode of evolution during the course of the search, to reach a satisfactory (or just interesting) solution.  The ability of a human user to apply powerful automated approaches like objective-based search~\citep{goldberg:gabook89, dejong:book02, eiben:book03, fogel:book06} and novelty search~\citep{lehman:alife08, lehman:ecj11} in short bursts and when appropriate is a key contribution of the NA-IEC approach.  The primary hypothesis is that letting the user make a relatively small number of critical selections during evolution, and leaving the remainder of search to automated approaches seeded by those user selections, can significantly augment the pace of evolution and the quality of its discoveries.

\begin{figure} [!t]
\centering
	\includegraphics[width=0.66\linewidth]{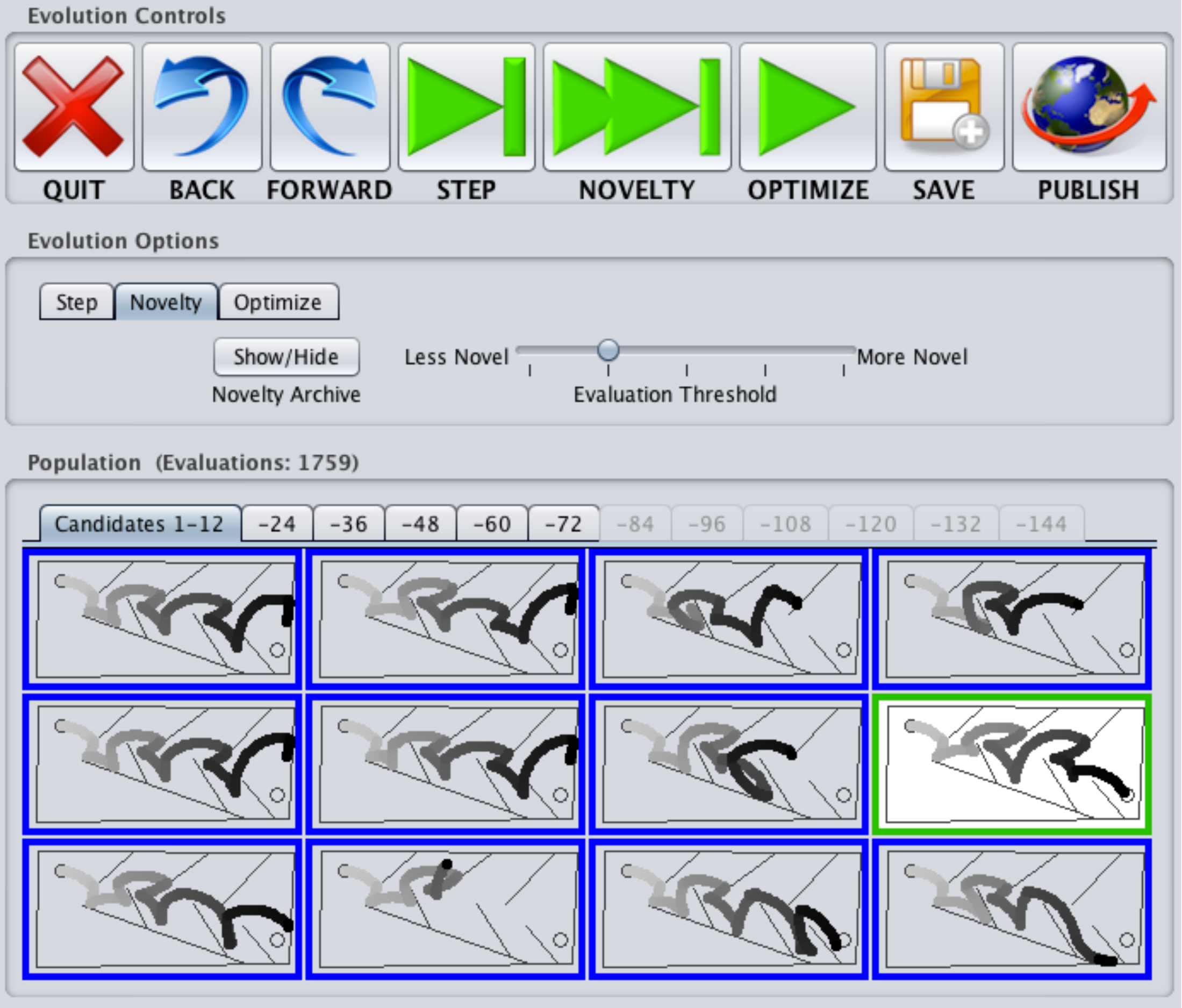}
	\caption{\textbf{Screenshot of the NA-IEC user interface.}  The user interface for the NA-IEC framework consists of the \emph{Evolution Controls}, the \emph{Evolution Options}, and the \emph{Evaluation Population}.  Candidate solutions are represented by a gradient trail that shows the robot's behavior in a particular maze.  Selected candidates are shown with a green border and solutions are highlighted with a white background.  Unlike traditional IEC applications, the user can now select one of three evolution modes: \emph{Step}, \emph{Novelty}, and \emph{Optimize}.  The \emph{Publish} button saves the results of a completed run for later analysis.  In the future it may be connected to the web.}
	\label{fig:naiecInterface}
\end{figure}

%TODO Describe a traditional IEC step
Figure~\ref{fig:naiecInterface} shows the main interface for the system, where the user can choose among the \emph{Step}, \emph{Novelty}, and \emph{Optimize} operations.  Choosing the \emph{Step} operation creates a new generation of offspring through the recombination and mutation of the selected candidate behaviors.  This classic approach to IEC is simple and computationally inexpensive, i.e.~it only creates a handful of new candidates.
 
%TODO Describe a Short-term Novelty Step
Choosing the \emph{Novelty} operation causes evolution to explore the space of agent behaviors without regard to an objective and then present the human evaluator with a broad view of where the evolutionary search can go from its current position.  To accomplish this aim, the next IEC population is generated by seeding a larger population with variations of the user-selected candidate behaviors and then running novelty search in the background to find novel individuals (in comparison to what has been encountered previously in the search) based on the sparseness measure $\rho(x)$ from equation~\ref{eq:noveltyMetric} and the threshold $\rho_{\text{min}}$.  The underlying evolutionary algorithm is NEAT~\citep{stanley:ec02, stanley:jair04}, which is often the base algorithm under novelty search~\citep{lehman:alife08, lehman:ecj11}.  Furthermore, to ensure that novelty is measured with respect to the entire search completed so far, all individuals encountered during \emph{both} traditional IEC steps and interleaved novelty searches throughout a session of NA-IEC are measured for their novelty and entered into the permeant archive if their novelty score is greater than the threshold~$\rho_{\text{min}}$.

The novelty search runs until \emph{at least} $n$ new individuals are added to the evaluation population (more than $n$ such novel candidates may be found when the novelty search is first started by generating an initial pool of candidates based on the user-selected choices on the screen), where $n$ is the size of an on-screen IEC population.  At that point, the collected novel individuals become the next IEC generation and control is returned to the user.  By convention, the $n$ novel individuals are sorted by their novelty score before the NEAT-based speciation adjustment to place the most novel candidate behaviors on the first visible page of the on-screen IEC population (figure~\ref{fig:naiecInterface}).  While the \emph{Novelty} operation is significantly more computationally expensive than the \emph{Step} operation, it provides the human user with a breadth of stepping stones that would have been time-consuming or impossible to discover on his or her own under the narrow view of a traditional single IEC step, which only presents the user with a handful of direct one-generation descendants.  In a sense, the set of stepping stones returned to the user by novelty search is like the set of images evolved by \emph{other} users from which a visitor to Picbreeder can branch:  In both cases, someone or something else has put in effort to collect a set of interesting jumping-off points and present them to the user.

By augmenting the human-led interactive search with interleaved novelty searches, a small population can be constructed that contains a set of novel stepping stones around the currently-selected candidates.  In the event that evolution cannot fill the next generation with a sufficient number of new archive members in a reasonable amount of time, the evaluation threshold can be decreased incrementally to allow the search to conclude quickly.

This approach does \emph{not} imply that the set of novel agent behaviors presented to the evaluator will be good at a potential task.  What is important is that they are \emph{behaviorally} diverse; it is the human evaluator who will direct the search by recognizing what is promising for a given domain.  The goal is to promote innovation through serendipitous discovery, and presenting the various directions that the search can take leverages the human evaluator's inherent ability to recognize what is important or interesting in a particular domain.

Finally, because objective-based optimization is likely the best option for \emph{perfecting} well-formed behaviors already discovered, the user is also given the option to request seeding a traditional objective-based search with currently-selected individuals.  The obj\-ec\-tive-based search will run until a specified solution criterion is met or until the user requests it to terminate, at which point the most fit individuals discovered so far will update the on-screen IEC population.  Providing this traditional option will allow users to optimize candidates that are near an objective attractor that the user would prefer to approach automatically once it is within striking distance (i.e.~once the search is no longer deceptive and the primary discovery is already made).

In this multifaceted approach, the user is free to change the mode of evolution between generations, thus allowing evolution to proceed in the capacity best suited for the current context.  In this way, the human user may begin NA-IEC by exploring the space of behaviors agnostically, and once an interesting behavior is established, the mode of evolution may be changed to optimize it.

%-------------------------------------
\section{Experiment} \label{sec:experiment}
%-------------------------------------
To demonstrate the synergistic effects of augmenting a human-directed search with novelty search, the experiment is conducted in the deceptive maze domain introduced by \citet{lehman:alife08, lehman:ecj11} (Section~\ref{sec:mazeDomain}).  That way, the NA-IEC approach can be compared against pure novelty search and fitness-based search directly.  In the deceptive maze domain, the goal is to evolve a navigation behavior that drives a robot from the start to the finish of the medium maze or the hard maze shown in figure~\ref{fig:maps}, which are constructed with several cul-de-sacs that create local optima in the fitness landscape.  Interestingly, these local optima are so deceptive that \citet{lehman:alife08, lehman:ecj11} found that novelty search significantly outperforms objective-based search in both mazes.  The question here is, can NA-IEC do even better?

To compare performance, each approach is evaluated over 30 runs on the medium and hard maps.  While novelty search and fitness-based search are both automated algorithms, the NA-IEC approach requires a human evaluator.  To accomplish the NA-IEC portion of this experiment, six users (who are not the authors) were recruited who were familiar with novelty search and EAs.  These users were introduced to the NA-IEC framework and each asked to evolve five solutions to the medium map and five solutions to the hard map.  The aim is to characterize the performance that can be reasonably expected from a practitioner in EC when evolving with NA-IEC.  Users were permitted to restart if they felt that evolution had become stuck.  However, all evaluations before such restarts were recorded as a part of the same run.

\begin{figure}
\centering
	\subfloat[Medium Map]{
		\includegraphics[width=0.338\linewidth]{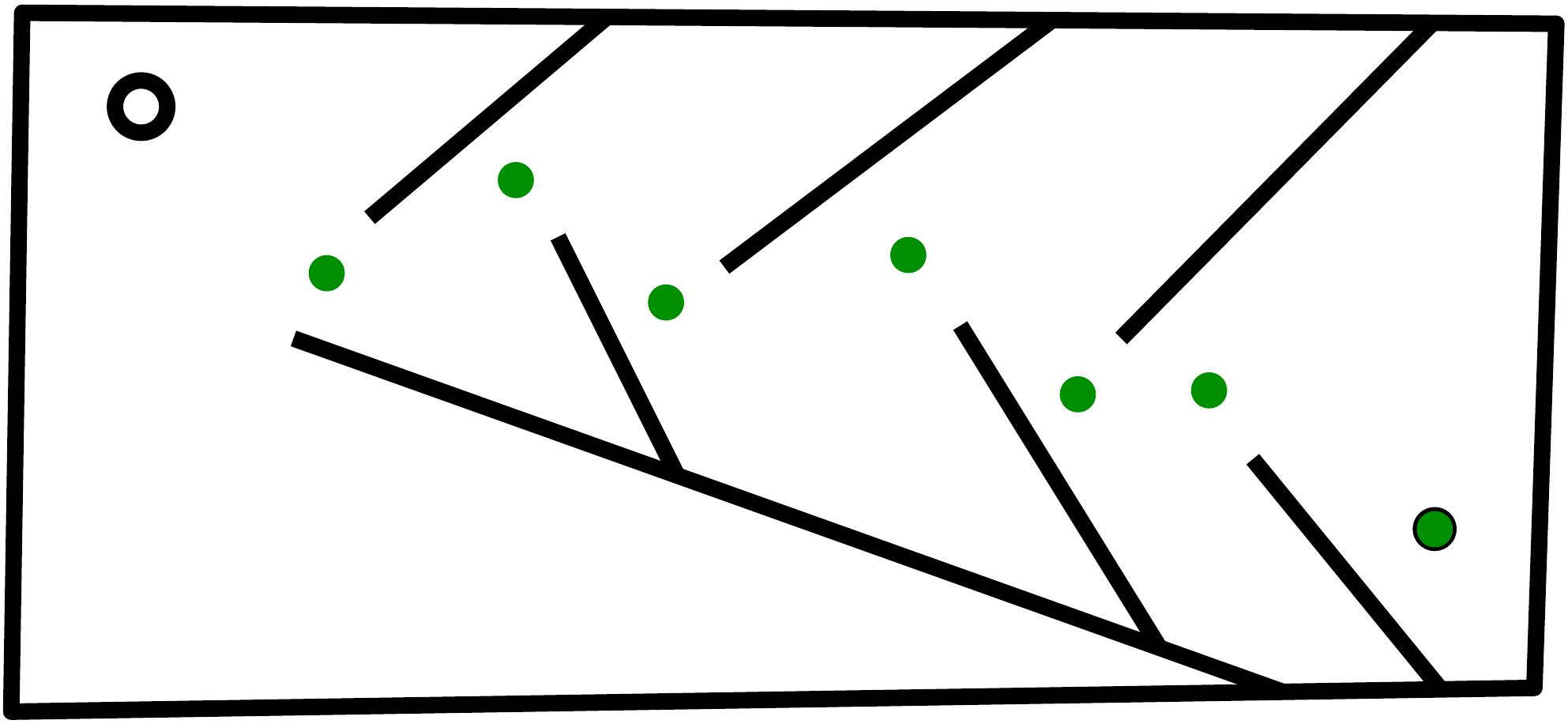}
	}
	\hspace{5em}
	\subfloat[Hard Map]{
		\includegraphics[width=0.225\linewidth]{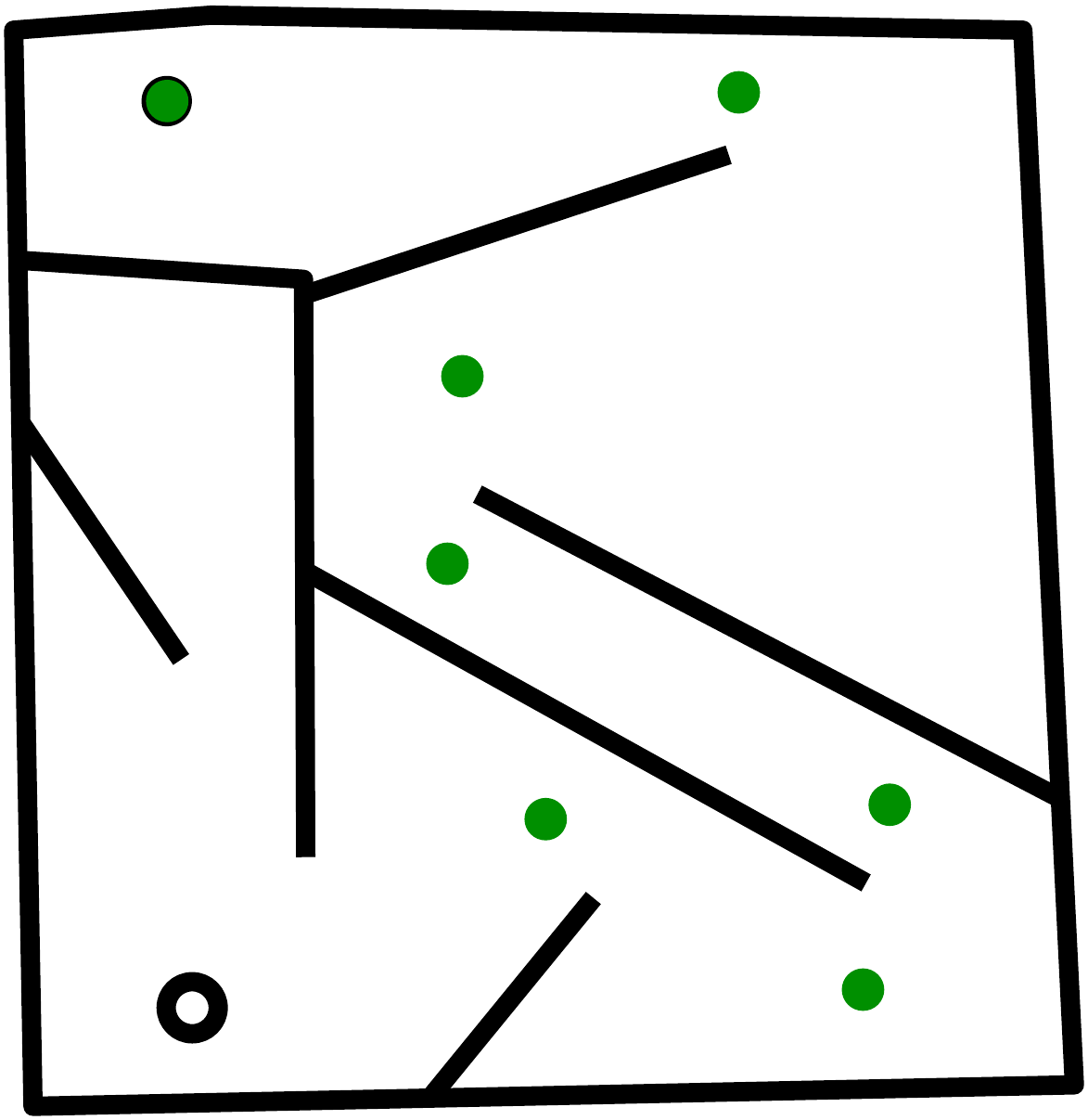}
	}
	\caption{\textbf{Maze navigation waypoints.}  To compare how much advantage is gained from knowing the path to the goal, waypoints (which are \emph{not} seen by the agent) are provided for the medium and hard maps.  In this way, deception is removed by allowing a traditional fitness-based search to reward solutions that discover stepping stones that are on the path to the goal.}
	\label{fig:waypoints}
\end{figure}

%-------------------------------------
%\subsection{Waypoint Directed Experiment}
%-------------------------------------
Inevitably, some will argue that such human guided runs have an unfair advantage because the user can see the path through the maze.  To address this concern, an additional fitness-based experiment, inspired by \citet{risi:gecco11} is conducted.  In this additional experiment the primary deceptive element of the maze navigation domain, i.e.~the attraction of agents to cul-de-sacs, is removed.  In this alternative reward scheme, candidates are rewarded for progressing along a path that actually leads to the goal.  Figure~\ref{fig:waypoints} shows the waypoints (which are invisible to the agent) in the medium and hard maps.  In this waypoint-directed version of the experiment, the fitness function $f$ is defined such that agents are rewarded for each waypoint crossed (including the goal); they also receive a partial reward for approaching the next waypoint:
\begin{equation}
	f = n + (1-d),
\end{equation}
where $n$ is the number of waypoints reached and $d$ is the distance to the next waypoint (proportional to distance between waypoint $w_{n}$ and $w_{n+1}$, in the range $[0, 1]$).

%-------------------------------------
\subsection{ANN Representation}
%-------------------------------------
In this experiment, as in \citet{lehman:alife08, lehman:ecj11}, the ANN controllers in all variants of the deceptive maze experiment are evolved by the NeuroEvolution of Augmenting Topologies (NEAT) approach~\citep{stanley:ec02,stanley:jair04}.  More specifically, the NEAT algorithm starts with a population of simple ANNs and \emph{complexifies} them over generations by adding new nodes and connections through structural mutations.  By evolving networks in this way, the topology of the network does not need to be known a~priori.  As applied to maze navigation policies, this process begins with an initial population of simple behaviors that are represented by fully-connected networks with 22 connections, no hidden nodes, and the inputs/outputs in figure~\ref{fig:robotANN}.  As the underlying networks add complexity (i.e.~new nodes and connections), features and nuances emerge in the resulting behaviors that could not be expressed by the simpler ANNs.

%-------------------------------------
\subsection{Experimental Parameters}
%-------------------------------------
The evolutionary parameters in this experiment are based on the deceptive maze navigation experiment by \citet{lehman:alife08, lehman:ecj11} and on the established parameters for NEAT~\citep{stanley:ec02}.  All experiments were run with a version of the public domain ANJI package~\citep{james:ANJI} augmented to support steady-state evolution, interactive evolution, and novelty search.  The IEC population size was 12, while the novelty search and fitness-based search population sizes were 250, with each run limited to 250,000 total evaluations.  Note that when the user initiates novelty search or optimization from within NA-IEC, a starting pool of 250 candidates are first generated from the user-selected candidates on the screen.  The speciation threshold, $\delta_t$, was 0.2 and the compatibility modifier was 0.3.  Recurrent connections within the ANN were allowed, offspring had a 5\% chance of adding a node, a 10\% chance of adding a link, a 1\% chance of loosing a link, and the weight mutation power was 0.8.  Unsigned activation was enforced in the ANN, resulting in a network output range that was shifted to $[-0.5, 0.5]$.  These parameters were found robust to moderate variation.

The parameters specific to novelty were also based on the original deceptive maze navigation experiment~\citep{lehman:alife08, lehman:ecj11}.  They include the nearest neighbors value ($k=15$) and the novelty threshold, $\rho_{\text{min}}$, which begins at 3.0 and is adjusted after every 2,500 evaluations.  Each navigation robot was given 400 timesteps to reach the goal, which only allows behaviors that proceed directly to the goal.  It is important to note that the experiments of \citet{lehman:alife08, lehman:ecj11} were \emph{re-run} with this setup to ensure a fair comparison and to validate our implementation.

%-------------------------------------
\section{Results}\label{sec:results}
%-------------------------------------
As with the original experiment by \citet{lehman:alife08, lehman:ecj11}, a navigation behavior that finishes within five units of the goal location is considered successful.  The main result is that NEAT with NA-IEC discovers such solutions in significantly fewer evaluations than both NEAT with novelty search and fitness-based NEAT on the medium and hard maps.  Furthermore, despite the expense of waiting on the human to evaluate a panel of candidate solutions, NA-IEC also consumes less clock time in search, suggesting that the value of the user's direction easily offsets the delay of waiting for human input.  Another result is that NA-IEC produces solutions with significantly fewer hidden nodes than both novelty search and fitness-based search, further suggesting the importance of allowing a human evaluator to make key decisions about the direction of evolution.  While some may dismiss such improvements based on the human evaluator's ability to see the path through the maze, results from the waypoint-directed search, a non-deceptive fitness-based experiment, are on par with NEAT with novelty search, which is still well below the performance of NEAT with NA-IEC.  The implication is that NEAT with NA-IEC not only exposes key stepping stones, but also provides evolution with subtle insights about the domain that are not easily incorporated into a traditional fitness function a~priori.

\begin{figure} [t!]
	\centering
%	\hspace{-1.85em}
	\subfloat[Medium Map]{
		\hspace{1em}
		\includegraphics[width=0.66\linewidth]{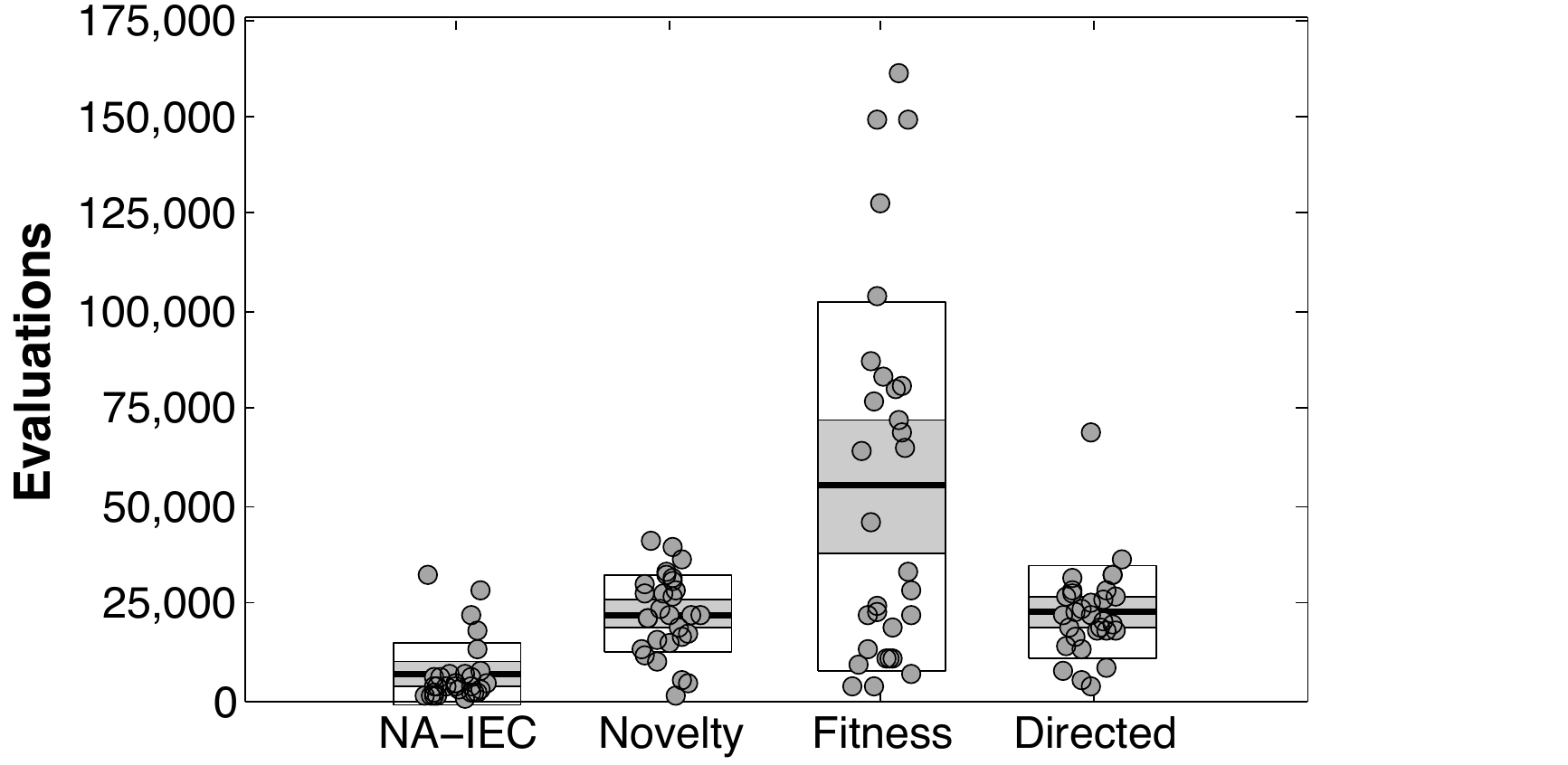}
	}
	
%	\hspace{-5.70em}
	\subfloat[Hard Map]{
		\hspace{1em}
		\includegraphics[width=0.66\linewidth]{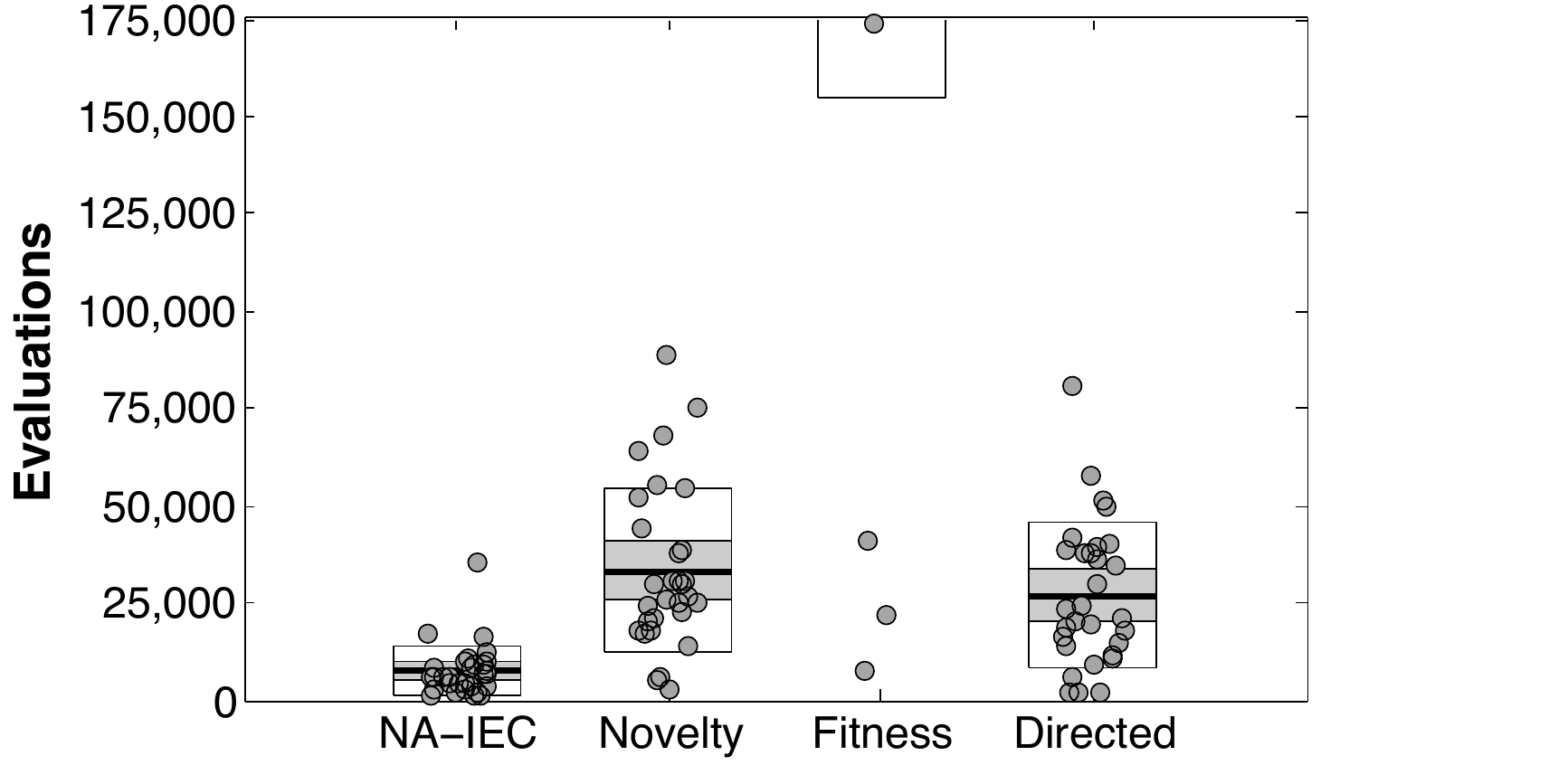}
	}
	\caption{\textbf{Evaluations required to find solutions.}  The number of evaluations required by NEAT with NA-IEC, NEAT with novelty search, fitness-based NEAT (pure), and waypoint-directed NEAT to find solutions are shown for the medium~(a) and hard~(b) maps.  The average number of evaluations to reach a solution is marked by a line while the boxed regions extend out to one and two standard-deviations; the distribution of the individual data points is also shown.  As in work by \citet{lehman:alife08, lehman:ecj11}, fitness-based NEAT is generally deceived in the hard map and is unlikely to produce solutions.  The main result is that the NA-IEC approach consistently finds solutions for the medium and hard maps in significantly fewer evaluations than not only novelty search and fitness-based search, but is also faster than fitness when the path through the maze is known.  Such results suggest that the human user's ability to recognize and select important characteristics as they emerge is directing evolution in a meaningful way.}
	\label{fig:performance}
\end{figure}

%\subsection{On the medium map...}
On the medium map, users directing NEAT with NA-IEC found 30 solutions in an average of 6,729 ($\text{sd}=8,068$) evaluations.  These results are significantly ($p<10^{-6}$; Student's t-test) faster than NEAT with novelty search (22,116 evaluations, $\text{sd} = 10,157$), fitness-based NEAT (55,066 evaluations $\text{sd} = 47,339$), and waypoint-directed NEAT (22,594 evaluations $\text{sd}=11,982$), each averaged over 30 runs (figure~\ref{fig:performance}a).  Furthermore, users solved the medium map in an average of 294 ($\text{sd}=359$) seconds, which is $2.8$ times faster than novelty search, $9.1$ times faster than fitness-based search, and $2.0$ times faster than the waypoint-directed search.  While solutions from novelty search, fitness-based, and waypoint-directed search have on average 3.2 ($\text{sd}=1.9$) hidden nodes, 2.9 ($\text{sd}=1.65$) hidden nodes, and 3.0 ($\text{sd}=1.8$) hidden nodes respectively, solutions produced by NA-IEC are significantly simpler, averaging just 0.23 ($\text{sd}=0.5$) hidden nodes per solution ($p<10^{-10}$; Student's t-test).

%\subsection{On the hard map...}
On the hard map, the NA-IEC approach evolved 30 successful navigators in an average of 7,481 ($\text{sd}=6,610$) evaluations, which is a significant ($p<10^{-5}$; Student's t-test) improvement over not only NEAT with novelty search alone (33,320 evaluations, $\text{sd} = 20,949$), but also over the non-deceptive (i.e.~waypoint-directed) version of fitness-based NEAT (26,954 evaluations, $\text{sd} = 18,464$), each averaged over 30 runs.  In the case of fitness-based NEAT, as in \citet{lehman:alife08, lehman:ecj11}, no comparison could be made because only four of 30 runs evolved solutions for the hard map.  Solution rates for the hard map are shown in figure~\ref{fig:performance}b.  In addition to evolving successful navigators for the hard map in fewer evaluations, NA-IEC did so on average in just 402 ($\text{sd}=374$) seconds, which is 3.5 times faster than NEAT with novelty search and $2.5$ times faster than the waypoint-directed search.  Regarding complexity, solutions from novelty search have on average 3.3 ($\text{sd}=1.8$) hidden nodes and solutions from the non-deceptive waypoint-directed search have an average of 3.5 ($\text{sd}=2.0$) hidden nodes, while those evolved by NA-IEC are significantly smaller with 0.5 ($\text{sd}=1.01$) hidden nodes ($p<10^{-8}$; Student's t-test).

\begin{figure} [p]
	\centering
	\subfloat[Medium Map Fitness]{
		\includegraphics[width=0.338\linewidth]{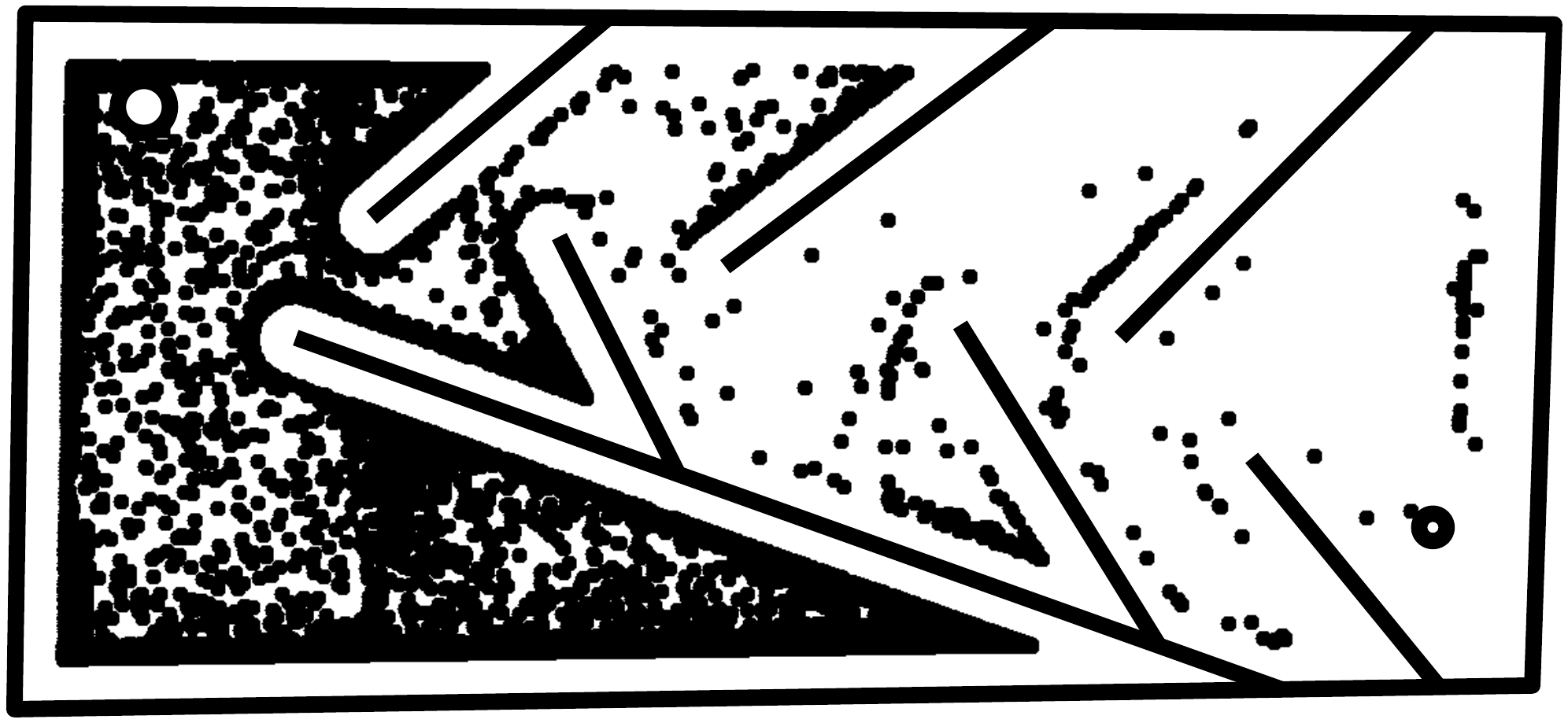}}
		\hspace{5em}
	\subfloat[Hard Map Fitness]{
		\includegraphics[width=0.375\linewidth]{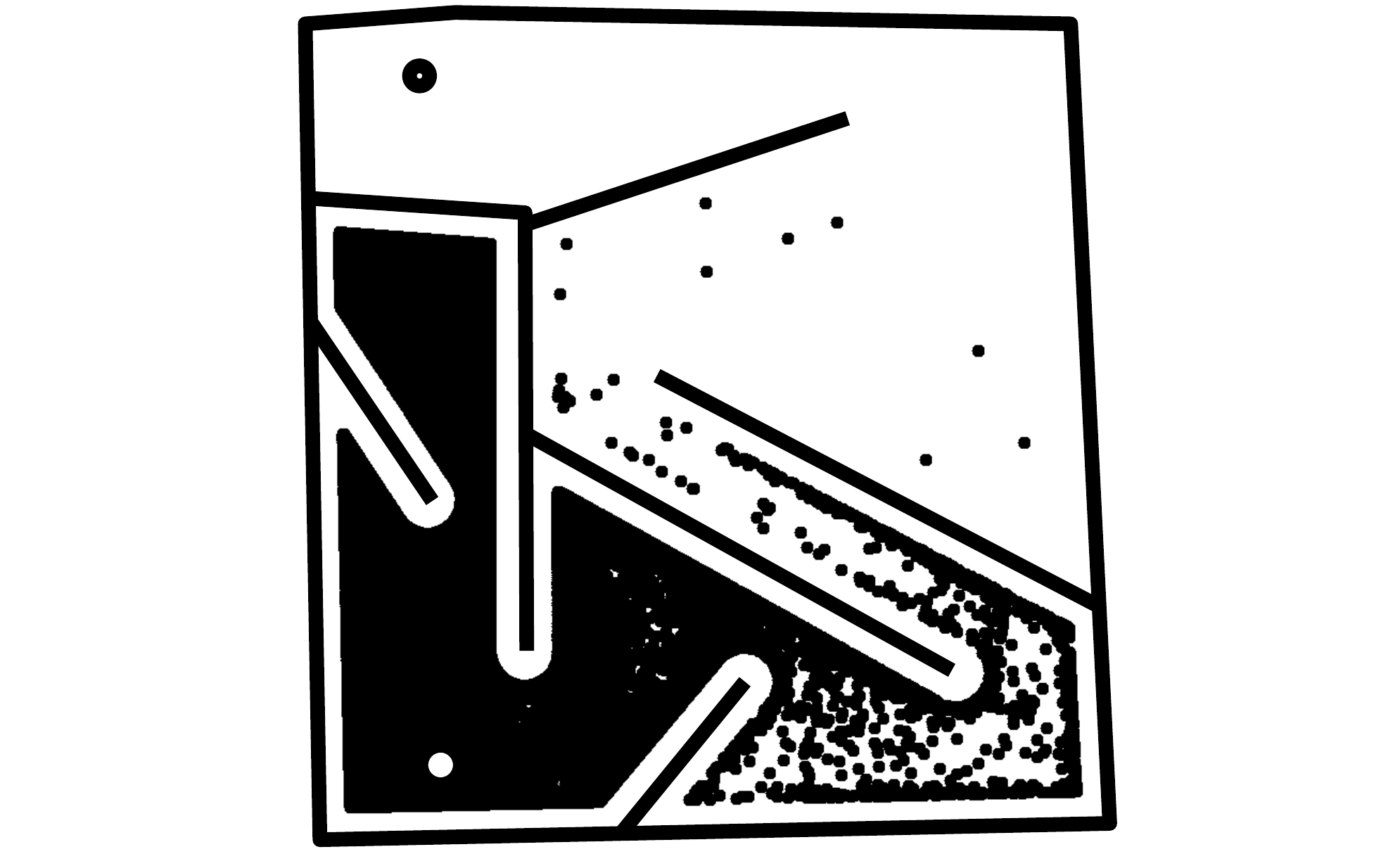}}
	\vspace{-0.5em}

	\subfloat[Medium Map Novelty]{
		\includegraphics[width=0.338\linewidth]{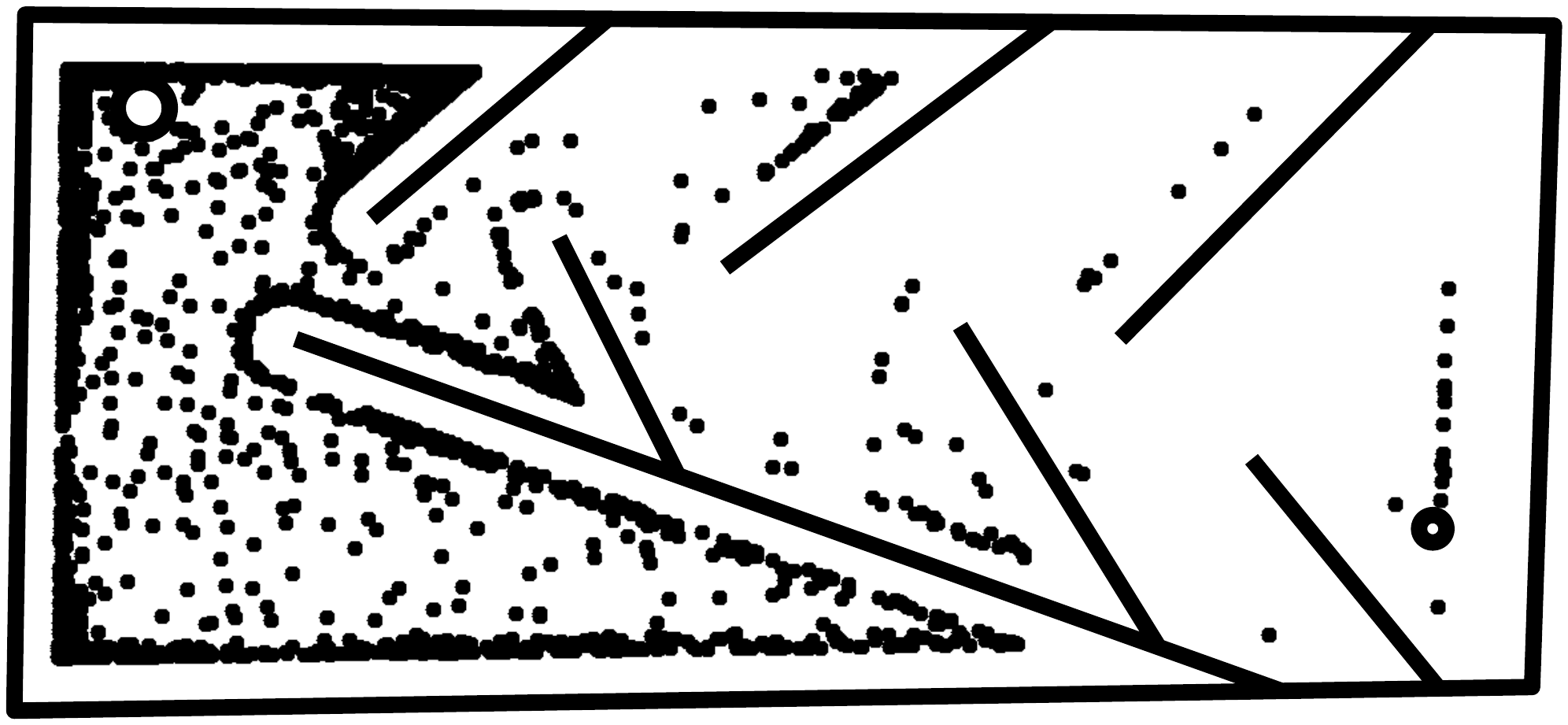}}
		\hspace{5em}
	\subfloat[Hard Map Novelty]{
		\includegraphics[width=0.375\linewidth]{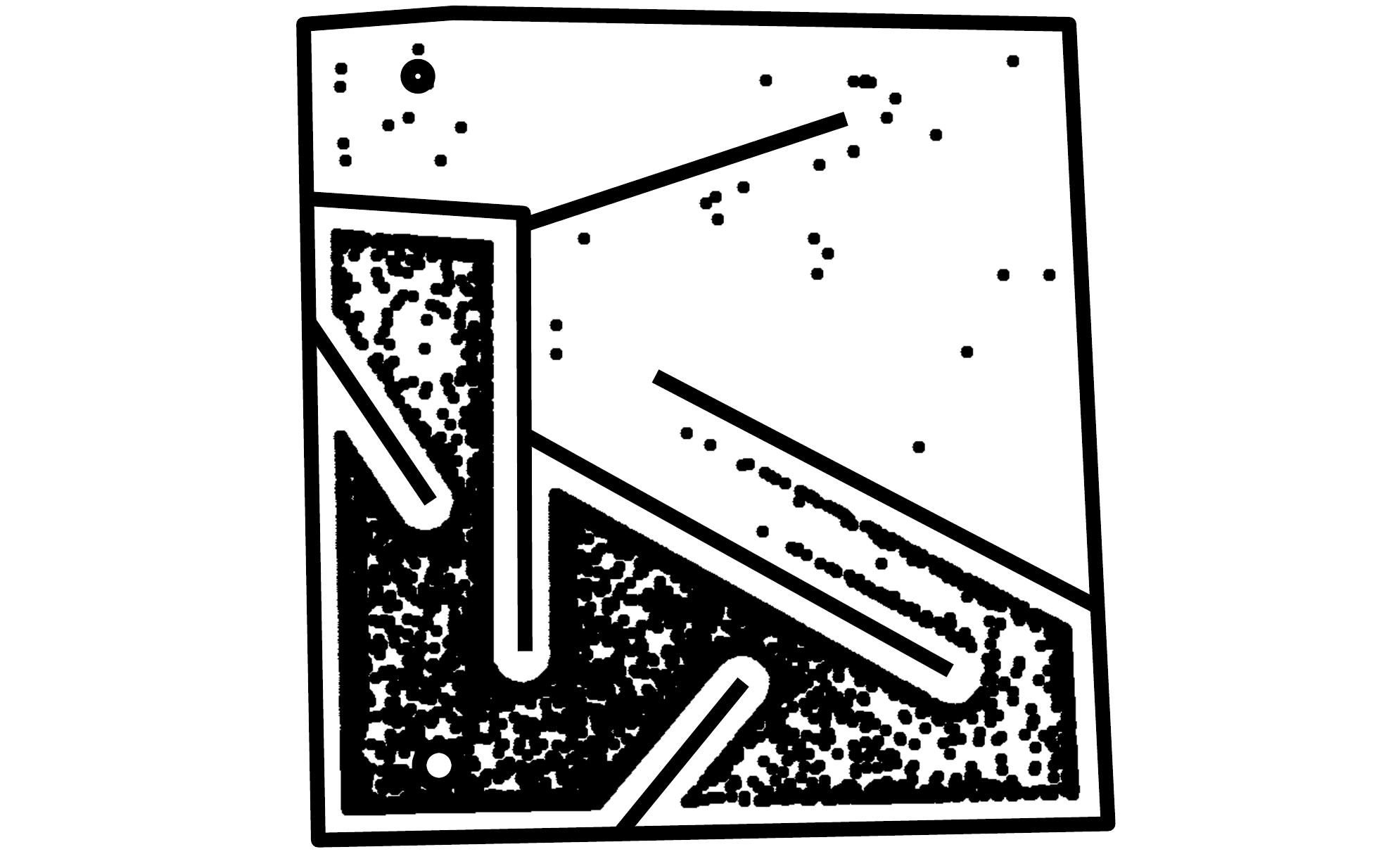}}
	\vspace{-0.5em}

	\subfloat[Medium Map Waypoint-Directed]{
		\includegraphics[width=0.338\linewidth]{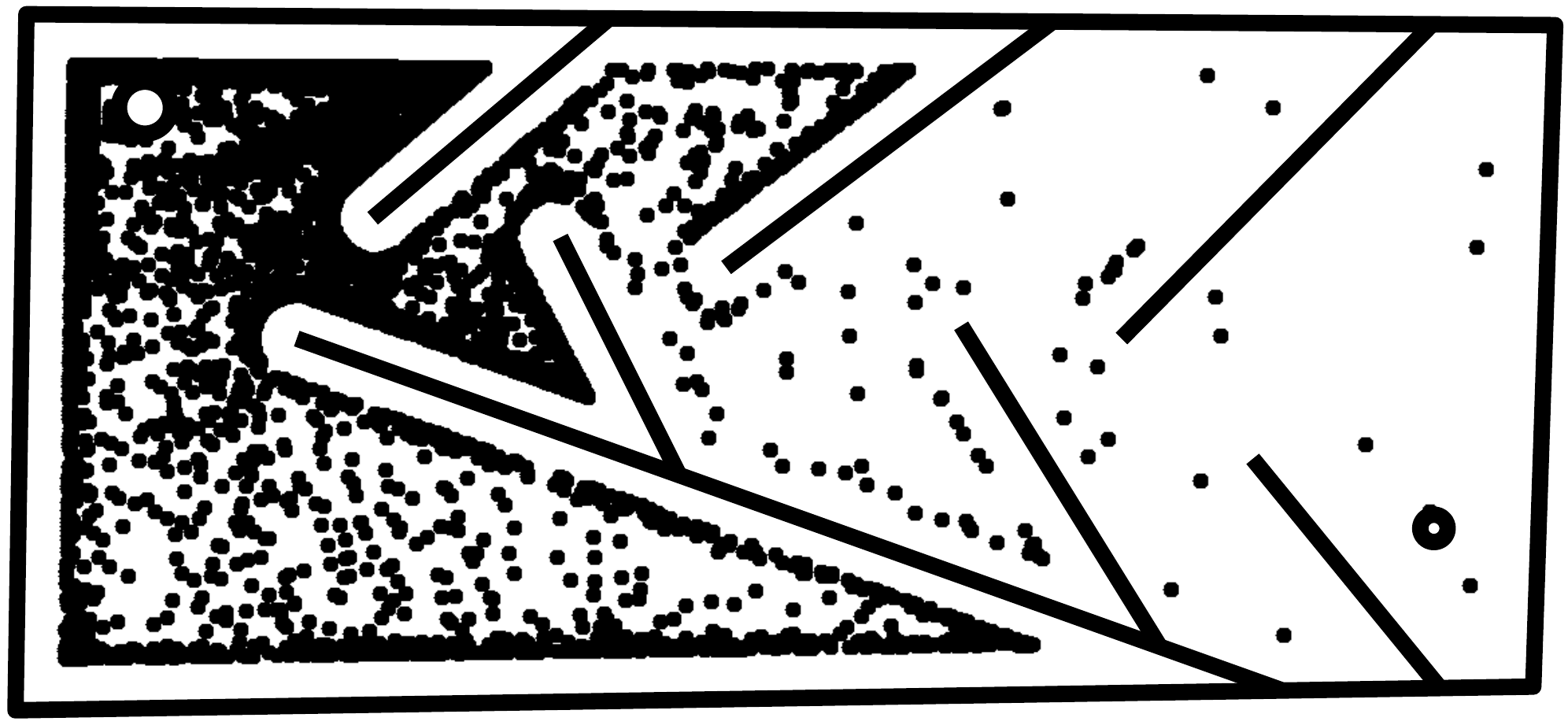}}
		\hspace{5em}
	\subfloat[Hard Map Waypoint-Directed]{
		\includegraphics[width=0.375\linewidth]{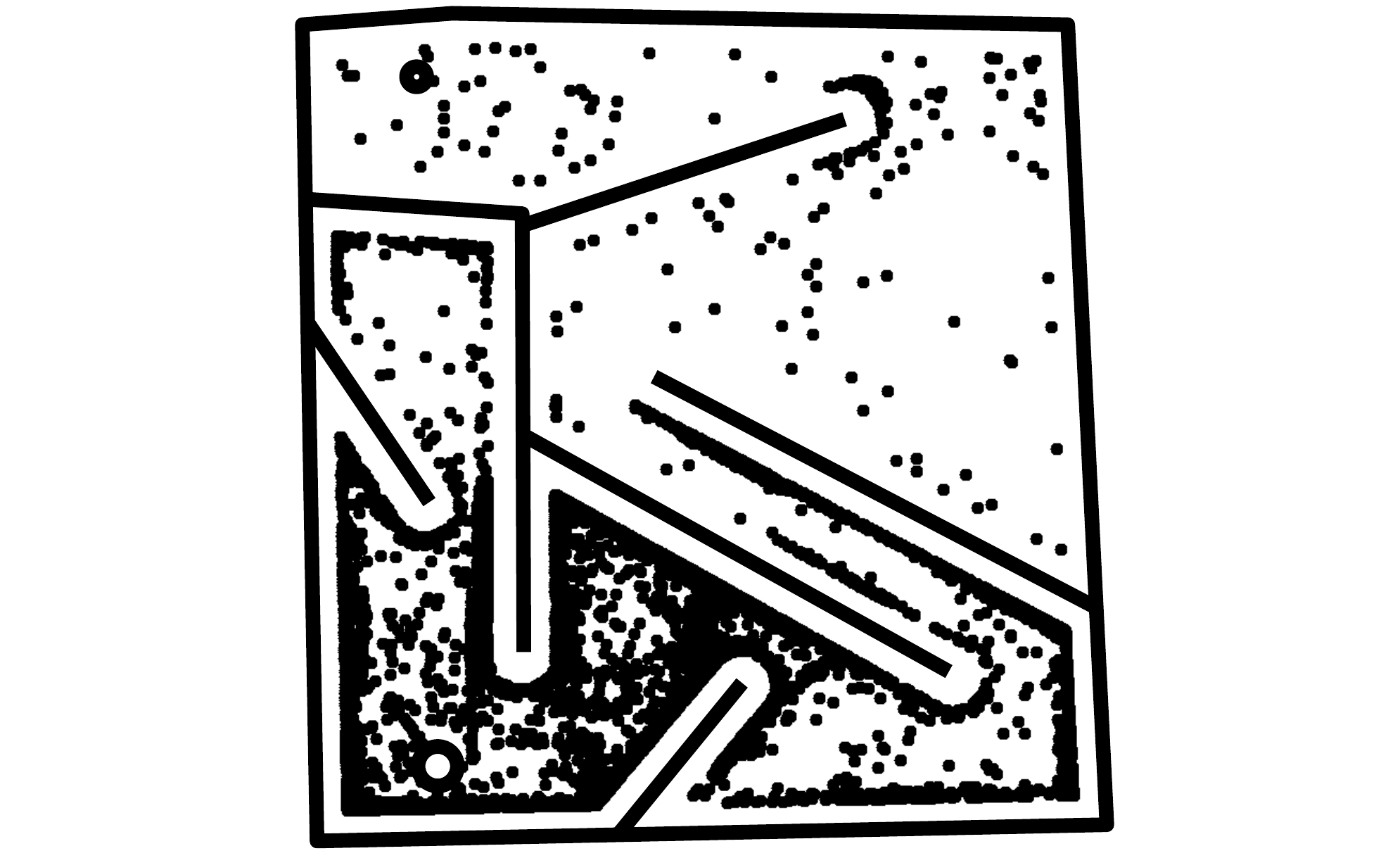}}
	\vspace{-0.5em}

	\subfloat[Medium Map NA-IEC]{
		\includegraphics[width=0.338\linewidth]{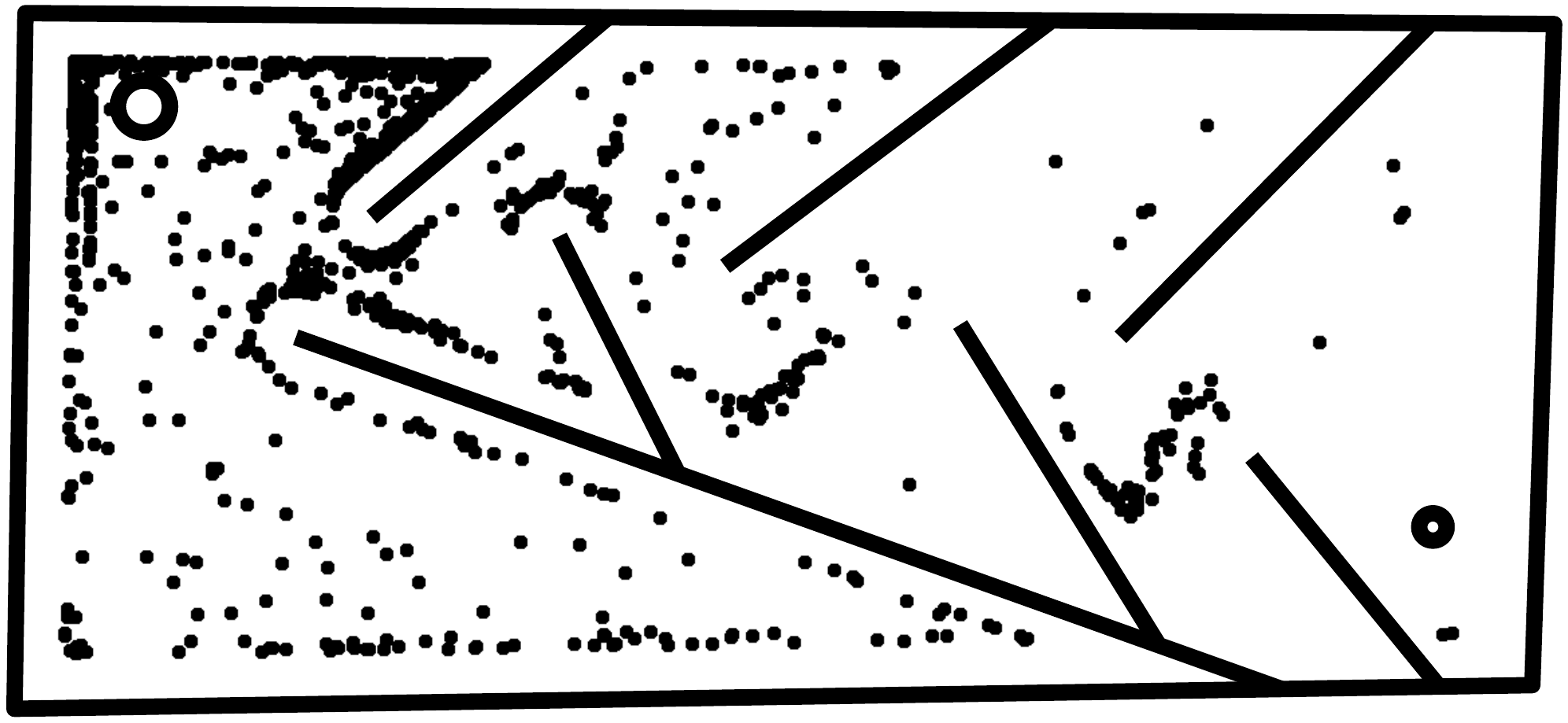}}
		\hspace{5em}
	\subfloat[Hard Map NA-IEC]{
		\includegraphics[width=0.375\linewidth]{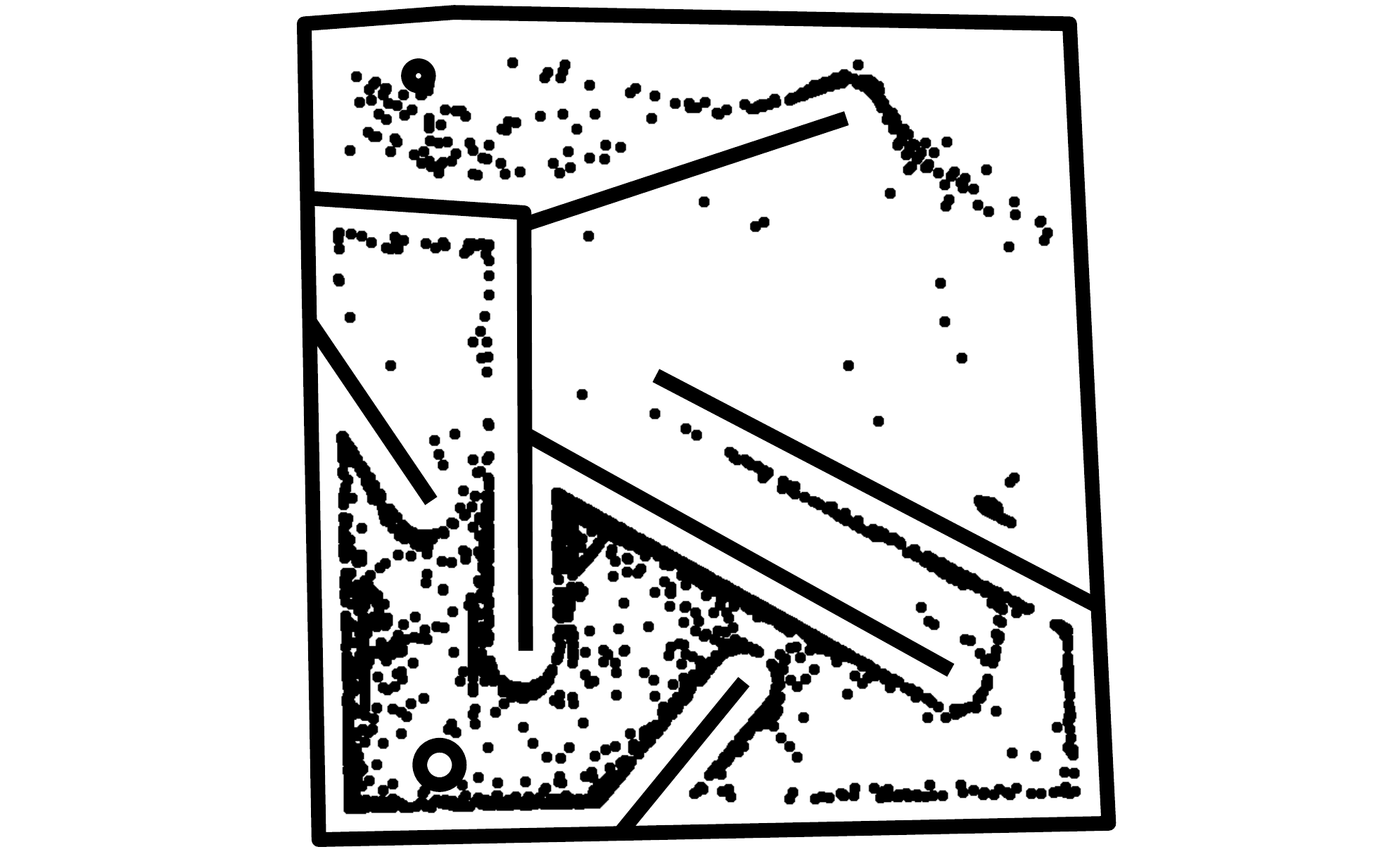}}
	\caption{\textbf{Distribution of final points visited.}  Each maze shows the final position for all candidates in a typical run.  The density of points shows how fitness-based NEAT, NEAT with novelty search, waypoint-directed NEAT, and NEAT with NA-IEC behave in the deceptive maze domain.  As in \citet{lehman:alife08, lehman:ecj11}, fitness-based search is attracted to the cul-de-sacs, while the points visited by novelty search are more evenly distributed throughout the maze.  When rewarded for progressing along the solution path, the waypoint-directed runs are less deceived.  For the NA-IEC runs, the human user's influence is clearly visible, i.e.~there are significantly fewer points in the major cul-de-sacs and tight groupings of points around key junctions.  Such characteristics reveal how human selections are impacting evolution.}
	\label{fig:behaviors}
\end{figure}

%\subsection{Typical Behavior}
Typical patterns of exploration for each approach in the medium and hard maps are shown in figure~\ref{fig:behaviors}, which compares the distribution of all ending points visited during a typical run.  As \citet{lehman:alife08, lehman:ecj11} discovered previously, the traditional fitness-based approach is attracted to the cul-de-sacs in the maze (figures~\ref{fig:behaviors}a and~\ref{fig:behaviors}b), while selecting for behavioral novelty allows NEAT to explore the space of possible behaviors more evenly (figures~\ref{fig:behaviors}c and~\ref{fig:behaviors}d).  Such search distributions are the result of selection pressure; thus when the objective-function rewards agents for following the solution path (figures~\ref{fig:behaviors}e and~\ref{fig:behaviors}f) the cul-de-sacs no longer deceive evolution.  Interestingly, when the points visited during NA-IEC are plotted in this way (figures~\ref{fig:behaviors}g and~\ref{fig:behaviors}h), the signatures of the human selector becomes evident.  As expected, the first of these is that there are far fewer points in the cul-de-sacs than in both novelty search and even the waypoint-directed search, demonstrating the intolerance of the human user for behaviors that explore these spaces.  The second signature is that there are frequently tight groupings of points at key junctions in the map, indicating that the user is probing these areas of the search space for a behavior that can turn a corner and enter a new chamber of the maze.  Such observations demonstrate how the human evaluator is contributing his or her insights to the search.  Furthermore, it is interesting how these human effects are so readily visible in the points plotted.

%\subsection{User Evaluations}
Finally, it is also important to analyze the behavior of the human users, especially in light of the human susceptibility to fatigue in IEC~\citep{takagi:ieee01}.  During the NA-IEC runs on the medium map, users made an average of 30.1 ($\text{sd}=40.5$) choices, applying the \emph{Step} function 29.8\% of the time, the \emph{Novelty} function 47.8\% of the time, and the \emph{Optimize} function 22.4\% of the time.  Similarly, solution were found for the hard map with an average of 32.0 ($\text{sd}=23.5$) human choices, of which 29.2\% were \emph{Step} functions, 58.9\% were \emph{Novelty} functions, and 11.9\% were \emph{Optimize} functions.  These statistics demonstrate that \emph{Novelty} is the preferred operation at most times, and that out of thousands of evaluations, only a few dozen user selections can dramatically reduce the overall cost of a run.

%-------------------------------------
\section{Discussion}
%-------------------------------------
In the deceptive maze domain, humans make a good team with novelty search and objective optimization, which helps to finish the job.  In both mazes, users choose \emph{Novelty} to generate the next set of choices significantly more frequently than the other options.  The stepping-stone generator of novelty search provides a desirable menu of possibilities to the human user, ultimately exceeding the performance of novelty search alone by several times.\footnote{The results in this paper also exceed the reported performance of novelty--fitness multi-objective hybrids~\citep{mouret:iros11}.}  Nevertheless, a natural question is whether such results are somehow specific to the maze domain.  Perhaps humans harbor a particularly keen insight into the most promising robot behavior in mazes, but would lack such insight in other domains.
 
For example, one hypothesis might be that humans in effect know the right path through the maze because they can see the whole maze.   Yet this interpretation is not entirely accurate.  The correct path \emph{through the maze} is not equivalent to the correct path \emph{through the search space}.  While some behaviors seem clearly dead ends (such as being caught in the most obvious cul-de-sac in the hard maze), others are less obvious.  It is not necessarily the case that just because one behavior drives the robot farther down the correct path that it must be a more promising stepping stone.  Some such behaviors are themselves dead ends that cannot push farther.  Also, humans perceive more subtle and nuanced indicators that are also important, such as path smoothness or unnecessary loops in the robot trajectory.   A behavior in which the robot doubles back on itself and then turns back onto the correct path may be just as ominous as being stuck in a dead end.  Humans intuitively understand \emph{these} kinds of dangers, yet to articulate them in an objective function would be quite challenging, and would almost certainly take more time than simply guiding the search away from them, whether they are easy to formalize or not.
 
In this sense, while only future empirical results can settle this issue, there is reason to believe that humans would carry similarly critical insights into other domains.  For example, in a biped-walking task~\citep{lehman:ecj11, reil:ieeetec02, vandepanne:ewas95}, humans can see that certain kinds of leg oscillations are promising even if the robot falls down.  Yet to describe exactly what makes them promising in a fitness function is likely prohibitive.  The human's overhead view, and hence knowledge of the mazes, should be viewed metaphorically as like any intuitive understanding of the shape of a particular behavior space.  Just as we can see in the maze that certain passageways must precede other passageways, so we can see in a biped robot that oscillations and balance must precede walking.  While it is possible that the intuitive insight into some domains is less than in the maze domain, the highly significant advantage provided by such insight in the maze domain suggests that even if the advantage were less elsewhere, it could still be significant. 
 
NA-IEC also may be important for more than just optimization.  In some spaces, such as in morphological evolution or with sophisticated encodings, we may be more interested in \emph{what is possible} than in achieving a particular end result.  The apparent synergy that results from humans combined with novelty search could be leveraged in the future to show us more about such spaces than trying to solve specific problems.  With all the limitations recently shown for objective-based search, NA-IEC provides an alternative without relinquishing our desire to have some say in the process, which is what the traditional fitness function usually facilitates anyway.
 
Finally, perhaps for some the involvement of a human will be unpalatable, violating a desire for total automation in machine learning.   Yet the human must be involved \emph{somewhere}.  After all, human researchers at the very least define the traditional fitness function for their experiments.  That is one reason NA-IEC was tested with humans with experience in EC.   Perhaps our effort and knowledge as researchers would be better applied by providing a modest set of hints to evolution that draw on our rich intuitive understanding of the domain, rather than through trying to articulate at the start of evolution an ad hoc formalization of what kind of behavior should necessarily precede what.  Humans in this study only spent up to ten minutes to make a few dozen selections among thousands of evaluations, much of which was automated by novelty search.  It is arguable that these few minutes represent time better spent than the time-consuming guesswork usually invested in crafting an objective function.  In any case, if our aim is to produce the very best results, as opposed to simply showing that an automated process can achieve a particular benchmark, then what we ultimately discover should matter more than how we get there.

%-------------------------------------
\section{Conclusion}
%-------------------------------------
This paper introduced the \emph{novelty-assisted interactive evolutionary computation} (NA-IEC) approach, wherein the intuitive ability of human users to identify promising stepping stones is augmented by an agnostic stepping stone generator (i.e.~novelty search) seeded with the behaviors selected by the user.  In this way, evolution proceeds unconstrained by a priori objectives, but still traverses key stepping stones that are meaningful to the human evaluator.  The result was a powerful synergy that allowed human users to realize what was important for a given domain \emph{during} evolution.  Furthermore, such serendipitous exploration found solutions in fewer evaluations, at lower genomic complexities, and in significantly less time overall than both novelty search and fitness-based search alone, suggesting that human direction in NA-IEC eases the need to craft domain-specific fitness functions.  Thus the key contribution of the NA-IEC approach is that it accelerates the rate and quality of evolution by leveraging human-level domain knowledge without burdening the user with the responsibility of evaluating every candidate created during evolution.

\end{document}